\documentclass[sn-chicago]{sn-jnl}% Chicago-based Humanities Reference Style
\usepackage{pdfpages}
\usepackage{mathdots}
\newcommand{\cites}[1]{\citeauthor{#1}'s (\citeyear{#1})}
\jyear{2023}%

\theoremstyle{thmstyleone}%
\usepackage{placeins}
\theoremstyle{thmstyletwo}%

\theoremstyle{thmstylethree}%
\renewenvironment{table}[1][]%
{\tableorg[#1]%
\tablebodyfont%
\renewcommand\footnotetext[2][]{{\removelastskip\vskip3pt%
\let\tablebodyfont\tablefootnotefont%
\hskip0pt\if!##1!\else{\smash{$^{##1}$}}\fi##2\par}}%
}{\endtableorg}

\begin{document}

\title[Stability of autonomous morphology]{The natural stability of autonomous morphology: How an attraction--repulsion dynamic emerges from paradigm cell filling}


\author*[1,2]{\fnm{Erich} \sur{Round}}\email{e.round@surrey.ac.uk}

\author[3]{\fnm{Louise} \sur{Esher}}\email{louise.esher@cnrs.fr}
%% \equalcont{These authors contributed equally to this work.}

\author[1]{\fnm{Sacha} \sur{Beniamine}}\email{s.beniamine@surrey.ac.uk}
%% \equalcont{These authors contributed equally to this work.}

\affil*[1]{
    \orgdiv{Surrey Morphology Group},
    \orgname{University of Surrey},
    \orgaddress{
        %% \street{Street},
        \city{Guildford},
        \postcode{GU2 7XH},
        %% \state{State},
        \country{United Kingdom}
        }
    }

\affil[2]{
    \orgdiv{School of Languages and Cultures},
    \orgname{University of Queensland},
    \orgaddress{
        %%\street{Street},
        \city{St Lucia},
        \postcode{4072},
        \state{QLD},
        \country{Australia}
        }
    }

\affil[3]{
    \orgdiv{LLACAN},
    \orgname{CNRS},
    \orgaddress{
        %% \street{Street},
        \city{Paris},
        \postcode{94801},
        %% \state{State},
        \country{France}
        }}


%% Should be 150-250 words

\abstract{Autonomous morphology, such as inflection class systems and paradigmatic distribution patterns, is widespread and diachronically resilient in natural language. Why this should be so has remained unclear given that autonomous morphology imposes learning costs, offers no clear benefit relative to its absence and could easily be removed by the analogical forces which are constantly reshaping it. Here we propose an explanation for the resilience of autonomous morphology, in terms of a diachronic dynamic of attraction and repulsion between morphomic categories, which emerges spontaneously from a simple paradigm cell filling process. Employing computational evolutionary models, our key innovation is to bring to light the role of `dissociative evidence', i.e., evidence for inflectional distinctiveness which a rational reasoner will have access to during analogical inference. Dissociative evidence creates a repulsion dynamic which prevents morphomic classes from collapsing together entirely, i.e., undergoing complete levelling. As we probe alternative models, we reveal the limits of conditional entropy as a measure for predictability in systems that are undergoing change. Finally, we demonstrate that autonomous morphology, far from being `unnatural' (e.g. \citealt{Aronoff1994}), is rather the natural (emergent) consequence of a natural (rational) process of inference applied to inflectional systems.}

%% 4-6 allowed:




\maketitle

\section{Introduction}\label{intro}

\textsc{Autonomous morphology} refers to linguistically significant generalisations---such as inflection classes or recurrent distributions of stem allomorphy---which systematically structure a language's morphology, yet are irrelevant to the rest of the grammatical system. The existence of autonomous morphology poses a serious challenge to any functionally-grounded theory of linguistics, since it is squarely maladaptive: its presence (as opposed to its absence) offers no functional advantage and can impose substantial learning costs during acquisition. While the initial appearance of autonomous morphology is straightforwardly attributable to accidents of diachrony, the substantive challenge is to explain why it then persists for millennia, often out-lasting other linguistic subsystems that are functionally well-motivated. Specifically, inflection classes and recurrent distributions of stem allomorphy are subject to constant diachronic change through inflectional analogy, and analogical changes \textit{could} quickly lead to their demise by levelling them out: so, if they are maladaptive and a clear path to their extinction exists, why does this path not prevail? Here we propose an answer. By modelling the historical dynamics of inflectional systems subject to iterated analogical changes, we show that, under the right conditions, a simple strategy for analogical reasoning leads to the emergence and persistence of inflection classes and recurrent distributions of stem allomorphy.

Of central importance to our proposal is the nature of synchronic analogical inference, since such inference functions as a diachronic source of novel inflectional patterns. As we shall see, small changes in assumptions about synchronic analogy may have ramifications for long-term diachrony that are significant. Synchronic analogy is constantly at work, because speakers of inflectional languages will rarely have encountered every last inflected form of lexemes that they know \citep{Chan2008,BonamiBeniamine2016,BoyeSchalchli2019}. Consequently, on occasion during production speakers will need to solve what \citet{AckermanBlevinsMalouf2009} term the `Paradigm Cell Filling Problem’ (henceforth PCFP) and produce inflected forms on the basis of inductive reasoning. Under such conditions it is important to ask: given a set of assumptions about how the PCFP is resolved, what are the predicted long-term implications for an inflectional system that incrementally changes under its influence? Most significantly, can we discover pathways by which known typological properties of inflectional systems arise emergently out of a cell filling process iterated over and over in the course of synchronic communication? By pursuing this line of enquiry, though it is not possible to prove that typological property $P$ \textit{necessarily} arises from synchronic process $S$, it is possible to establish that $S$ \textit{could} be the source of $P$, under the right conditions---a valuable form of scientific hypothesis generation which philosophers of science term `how-possibly' reasoning \citep{Persson2012}. Here, we apply how-possibly reasoning to furnish an new explanatory hypothesis for the stable diachronic persistence of autonomous morphology.

	As a starting point, we outline the current debate about the nature of autonomous morphology, including its putative computational function and its possible origin as an adaptive property within a gradual process of inflectional self-organisation. We highlight the challenge of explaining the consistently observed tendency for the mean conditional entropy of inflectional systems in natural language to remain stable at a low-yet-positive level, rather than falling to zero, and entertain the hypothesis that the emergence and persistence of autonomous morphology may be intrinsically linked to this tendency (Section \ref{autonomous}).
 
Computational modelling offers an accessible means to test such hypotheses by simulating the evolutionary trajectory of inflectional systems, including the emergence of system-level structural properties. We conduct a detailed review of the properties, insights and limitations of two early iterative models implementing a paradigm cell filling task \citep{AckermanMalouf2015,Esher2015a, Esher2015b,Esher2017}. Although designed to simulate gradual inflectional self-organisation, both models consistently evolve lexicons which are uniform and therefore lack the stable, structured variation that typifies autonomous morphology in inflectional systems of natural language. Via careful attention to the relationship between variation, interpredictability and entropy, we trace this outcome to a structural property of the models: a dynamic of preferential attraction in which lexemes and cells can only ever become more similar to others (Section \ref{existing}).

	Our next step is to enrich a replica of the existing models with additional, modulable parameters reflecting a more realistic view of linguistic input and speaker reasoning, such as Zipfian frequency weightings and recourse to narrower or wider samples. The marginal effects of these parameters on the overall evolutionary trajectory of the models---slowing, but never altering, the progression to total uniformity---confirm our insight that it is impossible for a system governed by a pure-attraction dynamic to stabilise with structured variation and low-but-positive entropy (Section \ref{family}).
 
However, models with a pure-attraction dynamic do not exhaust all possibilities, and upon closer inspection, other models may be better motivated. We show that rational reasoners can be expected to attend not only to `associative evidence’ based on lexical similarities, but also to `dissociative evidence’ based on differences. Incorporating dissociative evidence into our model introduces a repulsion dynamic which promotes divergence between cells and lexemes, alongside the existing attraction dynamic promoting similarity. The combination of these two dynamics promotes the emergence of stable inflectional organisation with low-yet-positive entropy, akin to that observed in natural language: pressure to coalesce pushes the system towards a lower number of variants, while pressure to disperse pushes the system to keep the remaining variants distinct (Section \ref{NE-model}).
Our model illustrates how structures corresponding to autonomous morphology can spontaneously (naturally) arise and subsequently persist as a form of inflectional self-organisation, via a simple (natural) inferential process that attends to both associative evidence and dissociative evidence. Coupled with the crosslinguistic prevalence and observed resilience of autonomous morphology, our results significantly erode the rationale for characterising autonomous morphology as unnatural: on the contrary, we contend that it is a fundamentally natural feature of human language. Our study further showcases the explanatory power and potential of computational evolutionary modelling (Sections \ref{discussion}, \ref{conclusions}).

%% ================

\section{Autonomous morphology in inflection}\label{autonomous}

Before proceeding to modelling, we first introduce the autonomous, inflectional morphological phenomena which will be the focus of our interest.

\subsection{Autonomy in synchrony and diachrony}\label{autonomous-language}

Originating with \citet{Aronoff1994}, the term `autonomous morphology’ reflects the insight that morphological systems often contain structures which cannot be reduced to phonology, syntax and semantics.\footnote{Of course, most linguistic structures can be reduced to others if one is sufficiently willing to tolerate unmotivated, disjunctive lists which figure suspiciously often in a grammatical description---for instance, the word class `noun' could be reduced to a disjunctive list of lexical items if one tolerates the list recurring in rule after rule. For a concrete example, see \citet{dechene2022sanskrit}, an analysis of Sanskrit nominal morphology without autonomous morphology, obtained by tolerating the appearance in grammatical rules of recurrent, unmotivated, disjunctive lists of morphosyntactic features.} Just like these other major components of the grammar, morphology exists independently from other components, while readily interacting with them and exhibiting interface phenomena \citep[see][]{Anderson2008,Anderson2011,Maiden2011a, Maiden2011b,Maiden2013,Maiden2018,Esher2013,Esher2017a,Esher2023,PatoONeill2013,Esher_O’Neill_2022}. Categories and properties of the morphology that are autonomous in this sense are often referred to as \textsc{morphomic}, following \citet{Aronoff1994}. At least three distinct subtypes of morphomic structure have been distinguished in the literature \citep{Round2015}: inflection classes or ‘rhizomorphomes’; recurrent paradigmatic distribution patterns or ‘metamorphomes’; and recurrent patterns in exponence, termed ‘meromorphomes’. We concentrate here on rhizomorphomes (inflection classes) and metamorphomes, which have received most discussion in the literature.

Typological surveys indicate that both rhizomorphomes and metamorphomes are ubiquitous and pervasive in natural language (\citealt{Bach2020,Herce2020}; see also \citealt{Esher2022a} for some instances of morphomic structure in agglutinating inflection). Moreover, there are cases in which the diachronic origins of metamorphomes or rhizomorphomes have been identified. Invariably these correspond either to a sound change, or to an analogical change which modifies an earlier system of metamorphomes or rhizomorphomes \citep{Maiden2009,Maiden2018,Esher2017a,Esher2020,Esher2022a,Enger2019a,Enger2019b,Bach2020,FeistPalancar2021}. Furthermore, such patterns are demonstrably persistent and productive in diachrony \citep{Maiden2001,Maiden2003,Maiden2005,Maiden2013,Maiden2016,Maiden2018,Enger2014,Enger2019a,Enger2019b,Enger2021,Esher2015c,Esher2017,FeistPalancar2021}. Diachronic studies demonstrate the ongoing resilience and productivity of both types of morphomic structure over time, in particular as a template to which additional lexemes can be drawn, through instances of morphological analogy \citep{Maiden2009,Maiden2011a,Maiden2011b,Maiden2012,Maiden2013,Maiden2016,Maiden2018,MaidenONeill2010,MaidenSmithGoldbachEtAl2011,CruschinaMaidenSmith2013,Enger2014,Enger2019a,Enger2019b,Enger2021,ONeill2014,ONeill2018,Esher2013,Esher2015a,Esher2015c,Esher2016,Esher2017,Esher2020,Esher2023,FeistPalancar2021}. Thus, in attested pathways of diachronic development, the overall structural principle of morphomic structure is perpetuated: there is ongoing fluctuation in the shape and incidence of individual distributional patterns, but this does not translate to the extinction of morphomic structure overall.

\subsection{Autonomous morphology and predictability}\label{autonomous-theory}

Autonomous morphological structures are most readily formalised within theories of the Word-and-Paradigm or Item-and-Pattern tradition \citep{Blevins2016,Stump2015}, which give prominence to the notion of inflectional paradigms, and to the relationships of predictability that hold within and between them.

In quantitative studies such as ours, relationships of predictability are typically expressed in terms of entropy \citep{Shannon1948}. For instance, the difficulty in guessing the contents of a single paradigm cell, $X$, can be expressed as its entropy, $H(X)$, which is zero in the case of complete predictability and positive otherwise.\footnote{For a cell $X$ with $n$ exponents $\{x_1, x_2, \ldots, x_n\}$ whose probabilities of occurrence are $\{\text{Pr}(x_1), \text{Pr}(x_2), \ldots, \text{Pr}(x_n)\}$, $H(X)$ is defined as: $H(X) = -\sum_{i=1}^n \text{Pr}(x_i) \log\text{Pr}(x_i)$.} Additional measures exist for expressing predictability that holds conditionally across cells: the predictability of $X$ given knowledge of another cell $Y$ (e.g., the predictability of the nominative plural given the nominative singular) is the `conditional entropy of $X$ given $Y$', $H(X \vert Y)$. Conditional entropy $H(X \vert Y)$ will be low in two situations which will become important below: either when the overall entropy of $X$ is simply low (i.e., the contents of $X$ are easy to guess even in the absence of other information) or when knowledge of $Y$ materially improves one's ability to guess $X$. The latter case is significant when attempting to account for how speakers solve the ‘Paradigm Cell Filling Problem’ (PCFP) \citep{AckermanBlevinsMalouf2009,BlevinsMilinEtAl2016}, a question which has attracted significant attention over the past decade \citep{AckermanMalouf2013,StumpFinkel2013,Bonami2014,BonamiBeniamine2015,BonamiBeniamine2016,BeniaminePHD,GuzmánNaranjo2018,CotterellKirovHuldenEtAl2019,ParkerSims2020,Pellegrini2020,Naranjo2020,LefevreElsnerSims2021,BeniamineBonamiLuis2021}.

A key contribution of these investigations has been to verify the existence of strong empirical tendencies in the structure of predictability within inflectional systems. Descriptive linguists have repeatedly noted that the observed complexity of individual inflectional class systems is highly constrained in terms of the number of distinct classes and distinct exponents available \citep{Carstairs1987,Carstairs1994,Plank1991}, and have interpreted this fact as revealing a general crosslinguistic principle of `paradigm economy' \citep{Carstairs1994}. Probing the possible origins of this principle forms the initial impetus for Ackerman and Malouf's 2015 experiment (Section \ref{existing}). Computational studies on predictability serve to quantify these observations: though languages exhibit considerable variation regarding the number of morphological features and feature values in their inflectional systems and regarding the entropy of individual paradigm cells, the mean conditional entropy of the system---that is, the mean of $H(X \vert Y)$ taken across all pairs of cells $\{X,Y\}$---clusters around values that are distinctly low (\citealt{AckermanMalouf2013}; see also \citealt{ParkerSims2020}).

Correspondingly, it has been suggested that morphomic structure has an essentially computational function. Since morphomes by definition involve recurrent patterning within paradigms, and since recurrence increases predictability, morphomic structures will contribute to curtailing the conditional entropy of the inflectional system \citep{CarstairsMcCarthy2010,Maiden2013,Maiden2018,Enger2014,Esher2015c}. A related hypothesis is that morphomic structure may therefore be adaptive, in which case it may arise and spread via self-organisation, given the tendency of cultural transmission to favour the development of linguistic systems that are more readily learnable---such as systems with high predictability \citep[see e.g.][]{Kirby2002,KirbyHurford2002,Esher2015a,Esher2015b,Esher2017,Esher2023}.

\subsection{The explanatory challenge of persistence}\label{autonomous-challenge}

Nevertheless, morphomic systems are not the only means to achieving low conditional entropy. An inflectional system can just as well have zero entropy if it lacks inflection classes and morphomic stem distributions entirely. This raises a challenging question. If language systems appear to favour low conditional entropy, and if self-organisation provides diachronic pathways to yet lower conditional entropy, why do languages not self-organise all the way to zero conditional entropy by levelling all morphomic structures? What causes the observed diachronic stability of low-yet-positive conditional entropy levels?

An explanation for why low, positive conditional entropy is diachronically stable should clarify why morphomes are so common across languages, and shed light on whether or not they are `natural'. Currently, the crosslinguistic ubiquity of morphomic structure contrasts with a continued tendency in the morphological literature to treat such structure as an unnatural, marginal phenomenon only to be recognised in the absence of other possible analyses.\footnote{Where a morphological structure is fully or partially isomorphic with any extramorphological correlate, it is not uncommon for this correlate to be treated as a necessary and sufficient motivation for the structure, excluding a morphomic analysis \citep{Anderson2008,Anderson2011,corbett2015,KoontzGarboden2016,BermudezOteroLuis2016}.} Indeed, the very existence of morphomic structure is regularly contested \citep{TangNevins2013,NevinsRodriguesTang2015,Bowern2015,BermudezOteroLuis2016,KoontzGarboden2016}. Even authors who describe empirically observable morphomic structures do not necessarily present these as a natural component of language. Striking examples include Aronoff’s description of morphology as `a disease, a pathology of language’ given that morphology is not necessary for communicative function \citep[413]{Aronoff1994} and the same author’s characterisation of morphomes as `unnatural kinds’ \citep{Aronoff2016} or even `morphological patterns that have become completely unhinged’ \citep[24]{Aronoff1994}; for further discussion of this view, see particularly \citet[223-226]{Blevins2016} and \citet[22]{Maiden2018}.

\subsection{Modelling change and emergent autonomy}\label{autonomous-modelling}

One approach to seeking answers for the nature of autonomous morphology is to model the evolution of inflectional systems over time. Computational, iterative models offer a means of testing whether system-level properties of interest, such as low entropy, can emerge over time from the accumulation of small changes. The hypothesis that gradual inflectional self-organisation might also give rise to morphomic structure with low conditional entropy explicitly prompted development of earlier iterative models \citep{AckermanMalouf2015,Esher2015a,Esher2015b,Esher2017} on which the models in our current study are based.

\subsection{Contributions of this paper}\label{contributions}

By means of computational simulations, we examine mechanisms and conditions which may be relevant to the emergence of morphomic structure. We show that although previous models can successfully derive the tendency for inflectional systems to reduce in conditional entropy, they also incorrectly predict that the tendency will run to its logical endpoint, with morphomic structure extinguished and durable stability attained only at zero entropy. When we introduce reasonable assumptions about analogical reasoning which have previously been overlooked, we find that inflectional systems not only gravitate towards low, positive conditional entropy but also rest there, without dropping to zero. Our findings support the view that inflectional systems possess an inherent potential for developing morphomic structure; furthermore, we show that such structure can emerge spontaneously and then persist. As such, we produce the strongest counterevidence to date against the view that autonomous morphological structure is `unnatural’, unexpected or inherently fleeting in human language.

\FloatBarrier
\section{Two existing iterative models}\label{existing}

The starting point for our own models is a seminal experiment designed by \citet{AckermanMalouf2015} to investigate the emergence of inflection class systems in a language evolution context: i.e., the progressive self-organisation of an initially unstructured system with high conditional entropy, into a more learnable system with low conditional entropy.\footnote{Although initially developed with reference to the evolution of the language faculty, the inherent assumptions, objects and mechanisms of \cites{AckermanMalouf2015} experiment are more congruent with historical change in `modern’ human language.}  The experiment, which we detail and evaluate in this section, is set up to test the hypothesis that a linguistic system which changes according to a simple historical process---the production of individual inflectional forms by analogical inference---will exhibit an overall evolutionary dynamic that spontaneously leads to a low mean conditional entropy for the system. The experiment takes the form of an iterative model carrying out a paradigm cell filling task, in which the exponent of a given cell in a given lexeme is treated as unknown, so that its form must be predicted by inductive inference based on other, known exponents. Although the individual changes within the model are consistent with what a single speaker might innovate, the model overall abstracts away from individual speakers. It is not agent-based and does not explicitly model transmission chains involving distinct participants. Likewise, although the model does represent an inflectional system which is transmitted with small modifications across time, it contains no explicit model of intergenerational learning or learning biases. While the simulations trace continuity and change in an inflectional system overall, they do so without representing individual instances of speakers' learning, production and perception within a speech community.

\subsection{Initial state of the model}\label{existing-initial}

\cites{AckermanMalouf2015} model is initialised with a lexicon in which paradigms are populated with randomly distributed exponents. The lexicon contains 100 lexemes, assumed to be of a single morphosyntactic word class, and each lexeme has 8 paradigm cells. A paradigm cell contains an `exponent', and each of the 8 cells in the system is associated with a set of 3 available exponents.\footnote{In our replica models, we associate each cell with a set of 5 exponents; the choice of 5 versus 3 does not affect the material processes of the model, but allows certain model behaviours to be more clearly distinguished.} Within Ackerman \& Malouf’s model, exponents are represented as simple indices of discrete categories; in a real language they would correspond to affixes, stem allomorphs, prosodic patterns or any combinations thereof that function distinctively in the inflectional system. At the outset of the experiment, each lexeme is populated with exponents: for each of its 8 cells, an exponent is randomly selected from the set of exponents associated with that cell. Since there are $3^8 = 6,561$ possible combinations of eight exponents, most of the 100 lexemes, if not all of them, will exhibit an inflectional pattern which is unique within the system, and thus instantiate their own unique inflection class.

\subsection{Inference}\label{existing-inference}

Each cycle of the model simulates a paradigm cell filling task in which the goal is to predict the exponent of one randomly selected cell for one randomly selected lexeme. For terminological clarity throughout the study, we will refer to the held-out (‘unknown’) exponent as the \textsc{focal} exponent, to its cell as the `focal cell’, and to its lexeme as the ‘focal lexeme’.

\begin{figure}[ht]
\caption[Paradigm cell filling mechanism]{Paradigm cell filling mechanism. Rows are lexemes, columns are cells; the focal cell of the focal lexeme, marked `?', is to be filled. (a) Select a pivot cell and examine its contents in all lexemes. (b) Evidence lexemes are those whose pivot cell matches that of the focal lexeme. (c) Examine the focal cell contents of the evidence lexemes. (d) Exponents score +1 for each token; select the highest-scoring exponent (in this case, \texttt{x}).}
\centering
\includegraphics[width=9cm]{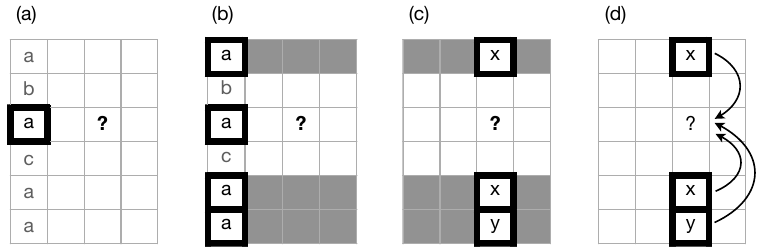}
\label{pcfp-positive-only}
\end{figure}

Prediction in each cycle is implemented according to the same defined strategy, which begins by randomly selecting an additional cell as a basis for inference; we will refer to this cell as the \textsc{pivot} cell. Next, the model notes the exponent of the pivot cell in the focal lexeme, as schematised in stage (a) of Figure~\ref{pcfp-positive-only}. The model then scans the pivot cells of all other lexemes, selecting those lexemes whose exponents in the pivot cell match the pivot exponent of the focal lexeme (Figure~\ref{pcfp-positive-only}, stage b). We refer to these selected lexemes as \textsc{evidence} lexemes. Next, the model notes the exponents of the focal cells of the evidence lexemes (Figure~\ref{pcfp-positive-only}, stage c). We will term these exponents `evidence exponents’. Among the evidence exponents, the model counts the number of tokens of each and selects the one with highest count. This exponent is the solution to the PCFP, and is copied into the focal cell (Figure~\ref{pcfp-positive-only}, stage d).

The last stage of the cycle evaluates the new version of the inflectional system. The new focal exponent is compared to the previous occupier of the same cell, noting whether or not the paradigm cell filling process has effected any actual change (as opposed to merely replacing an exponent with a copy of itself). As a tidying-up measure, the model also compares the focal lexeme to all other lexemes in the lexicon: if the focal lexeme is now identical to any other lexeme, the duplicate is deleted from the lexicon. By deleting duplicates, Ackerman \& Malouf's model maintains a one-to-one relationship between distinct inflection classes and `lexemes'. An alternative construal is that the model in fact only keeps track of inflection classes; for instance, when solving the PCFP, the choice of focal exponent from among the evidence exponents is based on a measurement of frequency in which each inflection class counts only once.

\subsection{Halting criterion}\label{existing-stopping}

The model continues to iterate until 25 cycles have passed without a change in the lexicon. The choice of 25 instances is arbitrary, and intended to represent the inflectional system reaching a somewhat stable state.

\subsection{Model outcomes}\label{existing-outcomes}

The simulation is repeated multiple times (it is implicit that the same initial lexicon is used for all simulation runs). The number of simulation runs is not stipulated, but instead is determined by the authors’ goal of collecting 500 examples of a stable state with more than one inflection class: in these 500 examples, the number of distinct inflection classes typically did not exceed 20. The authors further note that `in many cases [...] the final state has only a single inflection class, reflecting a complete leveling of the class differences in the starting state’ \citep[8]{AckermanMalouf2015}. However, they discard all such cases from their analysis: as a result, it is not known how many total simulation runs were made, nor how many runs ended in complete uniformity.

\citet[8,10]{AckermanMalouf2015} interpret the results of the experiment as demonstrating the spontaneous emergence of self-organisational principles in morphological systems, and it is certainly true that the experiment demonstrates a reduction in the mean conditional entropy of the inflectional system. It is significant that, in this model, low mean conditional entropy is not stipulated as a desideratum at any point (i.e. the evolutionary steps are not defined as ‘decrease mean conditional entropy’). Rather, low mean conditional entropy in the model is shown to be an emergent property when systems change according to the paradigm cell filling process described.

\subsection{Variations on the theme}\label{existing-variations}

Before discussing Ackerman \& Malouf's model further, we can note a number of modelling decisions which are made, but could also have been made otherwise.

Most obviously, the model could be run on inflectional systems of other sizes, with more or fewer lexemes, cells, and exponent options per cell; and the various cells could differ from one another in the numbers of exponents they make available.

Less trivially, the tidying-up process, in which duplicate lexemes are deleted, could be dispensed with. One noteworthy implication of tidying-up is that it prevents the model from ever increasing the overall number of inflection classes in the system: if each class $C$ has only one member and that one member is changed by the model to a new class $C'$, then its former class $C$ is entirely lost, whereas if $C$ has multiple member lexemes, then converting just one of them to $C'$ via the cell filling process still leaves class $C$ instantiated in the system, and the overall number of classes increases by one. Another is that tidying-up enforces uniform lexical type frequency for all classes, whereas if duplicate lexemes are retained rather than being deleted, then classes can increase in size. Where classes differ in size, larger classes will exert a greater influence on the outcome relative to smaller ones, due to their capacity to contribute more evidence lexemes at stages (b)-(d) in Figure~\ref{pcfp-positive-only}. In Sections \ref{family} and \ref{NE-model} we examine additional variants on this basic theme.

We can also mention an edge case, which we believe Ackerman \& Malouf's model was not intended to address. Namely, if the inflectional system is initialised so that the set of inflectional exponents available in each cell is similar in number to the total number of lexemes, then the paradigm cell filling mechanism behaves in a distinctive manner. Specifically, when the pivot cells of lexemes are compared (Figure~\ref{pcfp-positive-only}, stage b), the focal lexeme will often have no matches, or if it does have some, then only very few. This means that lexemes in the system interact with one another only very weakly; the lexicon is fractured into a large number of small, weakly-connected communities of lexemes. As the simulation progresses, lexemes within a community tend to become completely identical to one another, and moreover, lose all remaining similarities with lexemes outside their communities. As a result, the lexicon becomes partitioned into disconnected small groups of lexemes, with no inflectional exponents shared across groups. The outcome is that the system evolves multiple, stable inflection classes, but only because the structure was present from the beginning, in the almost-partitioned initial state: the evolution of the system amplifies this original structure rather than facilitating the emergence of novel structure.\footnote{To convey a sense of the magnitude of this effect: using 100 lexemes and 8 cells, we ran 500 iterations of the model (without tidying-up) initialised with each of 20, 40, 60, 80 and 90 distinct exponents available in each cell. With 20 exponents, none of 500 runs led to distinct stable classes; but with 40 exponents, 9 runs (approx. 2\%) did so; with 60 exponents, 21 runs did (approx. 4\%); with 80 exponents, 34 runs did (approx. 7\%); and with 90 exponents, 49 runs did (approx. 10\%).}
Our interpretation of \citet{AckermanMalouf2015} is that they were concerned not with edge cases such as these, in which the initial lexicon is characterised by near-partition-like structure, but rather with initial states that are unstructured. Similarly, here we set aside this edge case and only examine cases where the number of inflectional exponents in each cell is significantly lower than the total number of lexemes---as is true in the vast majority of the world's inflectional systems.

\subsection{Replication and critique of the model}\label{existing-replication}

Based on Ackerman \& Malouf’s description of their model, we constructed a replica and observed its behaviour over time; note that we removed the halting criterion and therefore allowed each model to run longer. Following \citet{AckermanMalouf2015}, we tracked the mean conditional entropy (mean $H(X\vert Y)$) and the number of classes throughout the simulations. In addition, we tracked the number of exponents per cell, as well as two novel measures: the average inter-predictability of cells (mean $U(X\vert Y)$) and a metric of class turnover, both introduced in more detail below. All measures are summarized in Table~\ref{table-measures}. 

\begin{figure}[ht]
\caption{Replication of \cites{AckermanMalouf2015} model, with tidying-up (i.e., deletion of duplicate lexemes after each change). Evolution of 100 initial lexemes with 8 cells and 5 exponents available in each cell. Black lines show mean values of 100 simulation runs. Grey ribbons indicate 90\% of runs' variation. All simulations ran for 2,500 cycles, indicated on horizontal axes. Note most vertical axes are non-linear, to enhance the visibility of model behaviour at low values.}
\centering
\includegraphics[width=12cm]{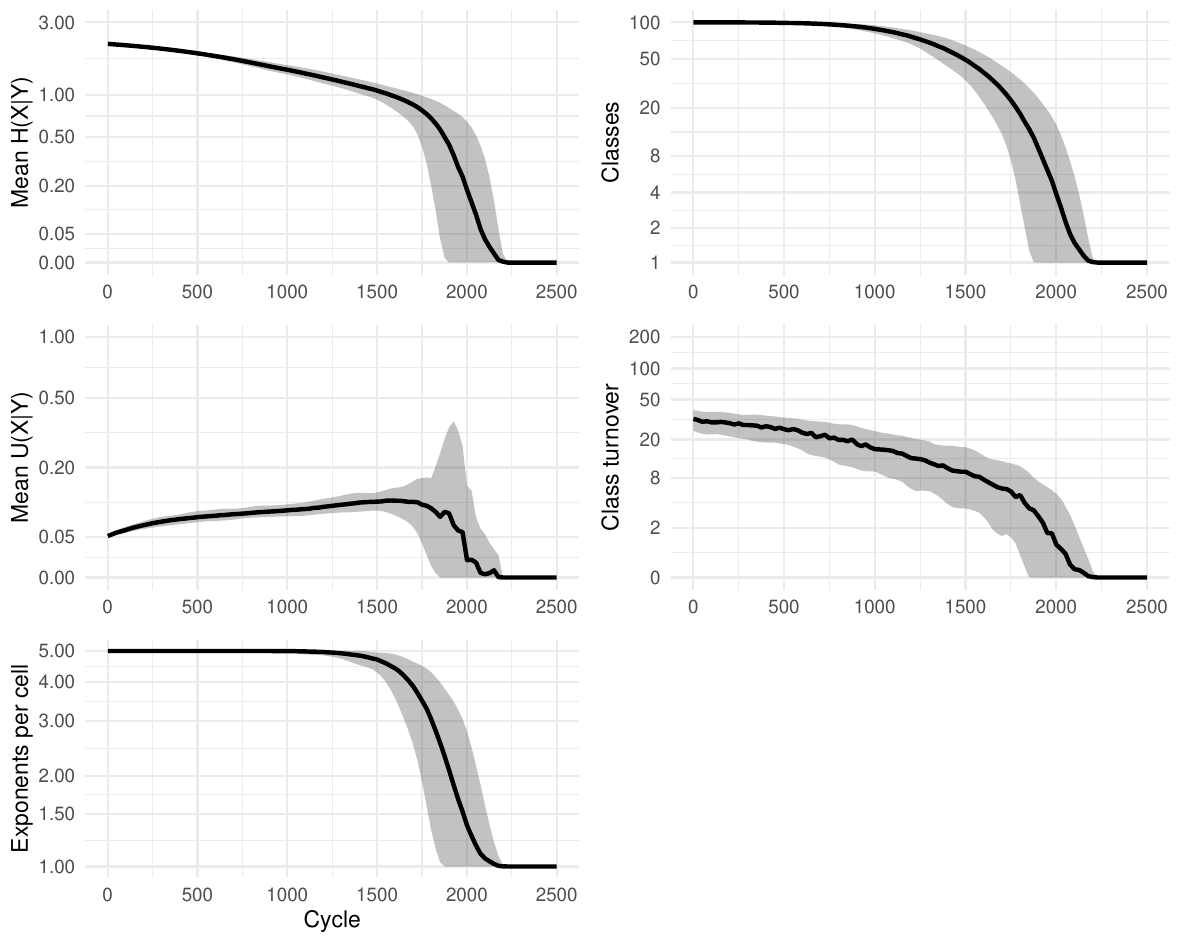}
\label{plot-AM-classwise}
\end{figure}

\begin{figure}[ht]
\caption{Replication of \cites{AckermanMalouf2015} model, without tidying-up (i.e., identically-inflected lexemes are tolerated, not deleted). 100 simulations for 10,000 cycles.}
\centering
\includegraphics[width=12cm]{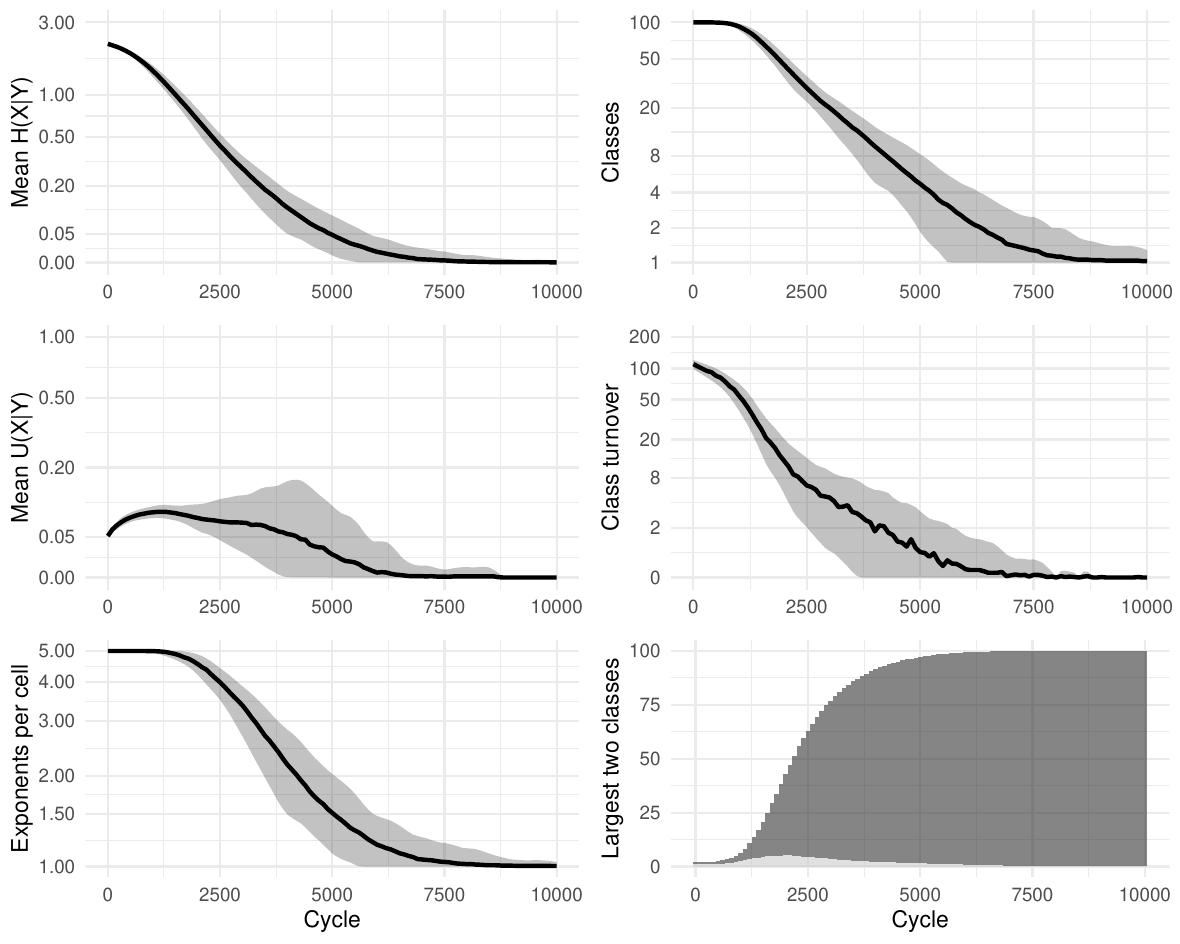}
\label{plot-AM-lexemewise}
\end{figure}

\begin{figure}[ht]
\caption{Eight snapshots evenly spaced between cycle 0 (leftmost) and cycle 10,000 (rightmost) from one simulation of \cites{AckermanMalouf2015} model without tidying-up. Each snapshot shows 100 lexemes in rows, 8 cells in columns. Distinct exponents in each cell are indicated by shading.}
\centering
\includegraphics[width=12cm]{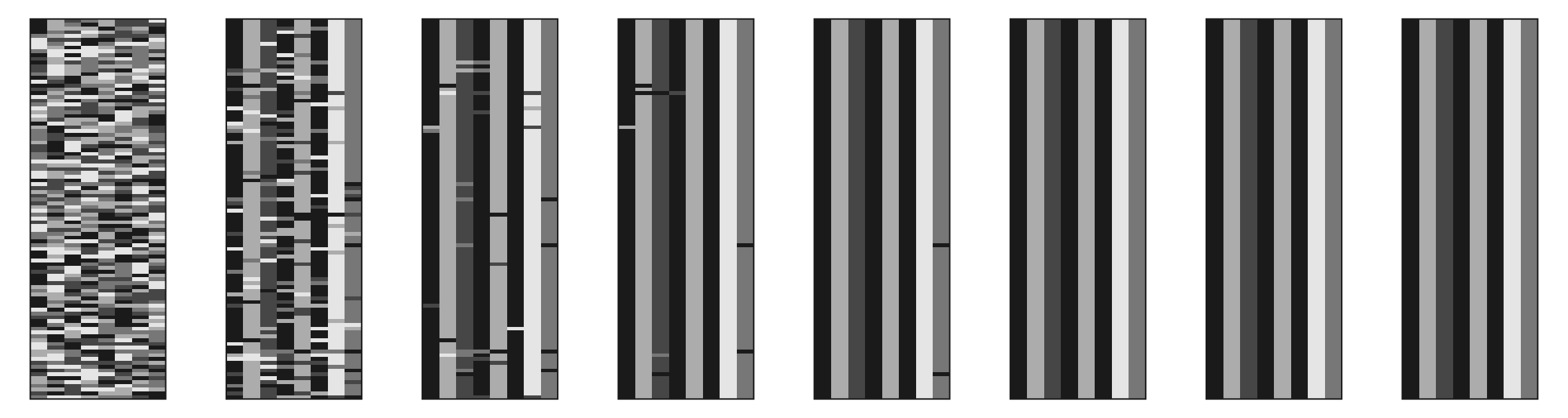}
\label{snapshots-AM-lexemewise}
\end{figure}

Our replica confirmed the apparent successes of the original model: namely, that the system self-organises, reducing the overall number of inflection classes, and that low mean conditional entropy emerges spontaneously. Figure~\ref{plot-AM-classwise} shows results from 100 simulations. We also ran a variant without tidying-up, allowing classes to grow in size beyond a single member. Essentially the same dynamics were observed, confirming that nothing in Ackerman \& Malouf’s original results is crucially dependent on the inclusion (or exclusion) of tidying-up. In all subsequent models, we dispense with tidying-up, thereby retaining all lexemes and allowing inflection classes to diverge in size. Figure~\ref{plot-AM-lexemewise} shows results from 100 simulations without tidying-up. Figure~\ref{snapshots-AM-lexemewise} shows snapshots of the inflectional system from eight evenly-spaced moments in one of the simulations, illustrating the progression of the lexicon from randomness to uniformity.

\begin{table}[!ht]
    \centering
    \caption{Measures of structure and complexity in simulated systems}
    \begin{tabular}{l|p{8cm}}
    \hline
        Measure & Definition and purpose \\ \hline
        Mean $H(X \vert Y)$ & Mean conditional entropy. Measures \textit{how predictable cells are} when another cell is known. Defined as the mean of $H(X \vert Y)$ across all pairs of cells $\{X,Y\}$. Takes a minimum value zero when any cell $X$ in the system is always entirely predictable, given any other cell, $Y$. Note that the value is influenced both by the independent predictability of cells $X$ before anything else is known, and by the increase in predictability gained by knowledge of other cells, $Y$. \\
        Mean $U(X \vert Y)$ & Mean Theil's U. Measures how cells \textit{improve each other's predictability}, that is, how much on average the predictability of any one cell $X$ improves with knowledge of another cell, $Y$. Defined as the mean of $\frac{H(X)-H(X \vert Y)}{H(X)}$ across all pairs of cells $\{X,Y\}$. Takes a minimum value zero when knowledge of a cell $Y$ always fails to increase the predictability of another cell $X$. Rises to a maximum of $1$ when every cell $X$ in the system is unpredictable without additional information but becomes entirely predictable given knowledge of any other cell, $Y$. \\
        Classes & Number of distinct inflection classes in the system. Minimum value of 1 corresponds to a lack of inflection class distinctions, hence complete predictability and zero entropy.  \\
        Exponents per cell & Mean number of distinct exponents available in each cell. Minimum value of 1 corresponds to no allomorpy, hence complete predictability and zero entropy.  \\
        Class turnover & A measure of the rate at which classes are innovated and lost, modelled on classic measures of species turnover in biology \citep{jaccard1901distribution}. Obtained by comparing two time slices $t_1, t_2$, and the sets of classes at those times $S_1, S_2$; defined as $\vert S_1 \cup S_2 \vert -  \vert S_1 \cap S_2 \vert$. Turnover is zero when the set of classes is unchanged between $t_1$ and $t_2$ and rises by 1 for each class present at one time slice but not the other. We set the gap between $t_1$ and $t_2$ as 1\% of the total simulation. \\
        Largest two classes & The size, in lexemes, of the largest inflection class in the system (dark grey) and the second largest (light grey). \\ \hline
    \end{tabular}
    \label{table-measures}
\end{table}

We tested models with other numbers of lexemes; cells; exponent options; and different numbers of exponents for different cells. Consistently, the same essential results are obtained: all systems evolve to uniformity. Thus, whereas Ackerman \& Malouf's original study placed some emphasis on the kinds of multi-class inflection systems that could be observed using their halting criterion (\citeyear{AckermanMalouf2015}:8,10), we suggest that these results are somewhat like dropping a ball and halting observations before it hits the floor. It is true that the ball has fallen (cf decreased in mean conditional entropy) but the fact that it `ends' several centimetres above the floor (cf retains more than one inflection class) is an artefact of the halting criterion; if observations continue beyond this point, the ball will always be found to continue its descent and reach the floor (cf reach zero entropy). So while \citeauthor{AckermanMalouf2015}'s model does reproduce a spontaneous progression to low conditional entropy which seems characteristic of real languages, it is unable to remain stable with a low-but-positive conditional entropy, i.e., with persistent morphomic structures in the form of multiple inflection classes.\footnote{If the prediction strategy is modified so that at Figure~\ref{pcfp-positive-only} step (d) it picks an exponent at random rather than the most frequent exponent, the result is a perpetually unstable system in which neither mean conditional entropy nor the number of inflection classes reduces at all \citep[11]{AckermanMalouf2015}.}

\begin{figure}[ht]
\caption{Diagrammatic representation of (a) entropy (circles) and conditional entropy (in grey), and two scenarios in which conditional entropy declines: (b) due to increased overlap, (c) due to decreased overall entropy.}
\centering
\includegraphics[width=11cm]{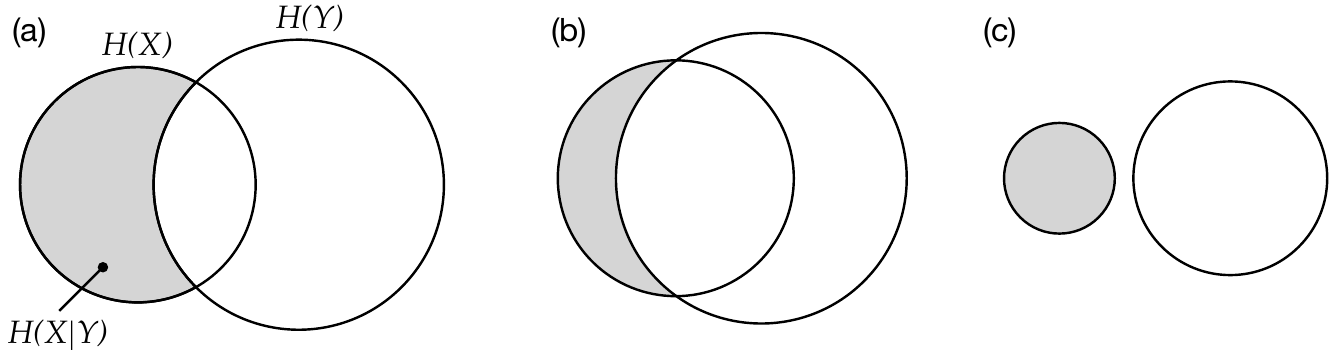}
\label{entropy}
\end{figure}

In addition to conditional entropy (presented in Section~\ref{autonomous-theory}), we introduce a new information-theoretic measure (Theil's $U$, \citealt{theil1972statistical}) in order to observe the evolving nature of predictability between cells. To appreciate the relationship between predictability and conditional entropy, it will be helpful to begin by returning momentarily to conditional entropy itself and how this property can change over time. In Figure~\ref{entropy}a, the size of the circle marked $H(X)$ `entropy of X' represents the uncertainty when guessing the exponent of a single cell, $X$; the circle $H(Y)$ represents the uncertainty when guessing the exponent of another single cell, $Y$. The zone marked $H(X  \vert Y)$, i.e., the part of $H(X)$ which does not overlap with $H(Y)$, is the uncertainty about cell $X$ which remains after we know the identity of the exponent in cell $Y$. $H(X \vert Y)$ is the conditional entropy of $X$ given $Y$.

To appreciate how these relationships can change over time, we invite the reader to picture, firstly, the situation in which the two circles in Figure~\ref{entropy}a move closer together over time so that the overlap between $H(X)$ and $H(Y)$ grows, as in Figure~\ref{entropy}b. In this scenario, as time progresses, cell $Y$ provides ever more information about $X$, thereby shrinking the area of $H(X \vert Y)$, and conditional entropy $H(X \vert Y)$ \textit{declines over time}. We suspect that \citeauthor{AckermanMalouf2015}'s experiment has been widely interpreted as conforming to this scenario: $Y$ becomes more informative about $X$ and for this reason conditional entropy declines. However, picture next the situation in which the circles in Figure~\ref{entropy}a stay in place but both shrink, as in Figure~\ref{entropy}c. In this scenario, the overlap between the circles would progressively decrease, and, as they continue to shrink, a point would be reached where there ceases to be any overlap at all. Notice that from the moment when overlap ceases, $H(X \vert Y)$ will simply be equal to $H(X)$. As the circles shrink further, $H(X)$ falls and so too does $H(X \vert Y)$. This is a scenario in which conditional entropy \textit{also declines over time}, but not because of increasing information sharing between cells $X$ and $Y$; on the contrary, the mutual predictability of $X$ and $Y$ collapses to zero. In Ackerman \& Malouf's model, a key factor that causes $H(X)$, for any cell $X$, to decline (and thus, the circle in Figure~\ref{entropy} to shrink) is the loss of variation of exponents in the cell. In every run of the model, as the number of exponents attested for a given cell invariably drops from 5 to 4, 3, 2 and ultimately just 1, $H(X)$ will also drop, eventually reaching zero. The two scenarios discussed here are crucially different, but conditional entropy alone is incapable of distinguishing between them.

In order to discern why conditional entropy is declining, a second metric is needed. Theil's U \citep{theil1972statistical}, written $U(X \vert Y)$, is defined as $\frac{H(X)-H(X \vert Y)}{H(X)}$, or zero when $H(X)=H(X \vert Y)=0$. It quantifies the degree to which knowledge of $Y$ enhances the predictability of $X$, cast as a \textit{proportion} of how much unpredictability $X$ has overall. In scenarios of increasing overlap between $H(X)$ and $H(Y)$, the value of $U(X \vert Y)$ will \textit{increase}, eventually reaching a maximum of $1$ when the two circles overlap and knowledge of $Y$ allows complete prediction of $X$. In contrast, in scenarios of shrinking $H(X)$ and $H(Y)$ leading to diminishing overlap, $U(X \vert Y)$ will \textit{decrease}, reaching a minimum of $0$ when the two circles no longer overlap at all, i.e., when knowledge of $Y$ no longer provides any information about $X$. In this way, $U(X \vert Y)$ successfully differentiates between the two different scenarios which conditional entropy conflates. Accordingly, we take measurements of $U(X \vert Y)$ to understand which dynamic is at the root of decreasing conditional entropy.\footnote{\citet{sims2023analogy} use Thiel's $U$, with the label `normalised mutual information', for comparing conditional entropies between different languages or between different analyses of the same language. As in our study, $U(X \vert Y)$ is useful because it produces meaningful comparisons in conditions where $H(X)$ is not held constant.}

In the simulations, we observe that at first, mean $U(X \vert Y)$ climbs briefly. Then,  as the system begins to lose exponents (as seen in the plots of `mean exponents per cell' in Figures \ref{plot-AM-classwise} and \ref{plot-AM-lexemewise}), mean $U(X \vert Y)$ slows its climb, reverses course and falls to zero, indicating a deterioration then complete collapse in the degree to which cells are improving each other's predictability. Although the model causes mean conditional entropy to drop, it does so not because cells are mutually reinforcing each other's predictability, but rather because cells are losing their internal variability.

We conclude that Ackerman \& Malouf's model has an unavoidable bias towards system uniformity. The ultimate reason for such bias is readily apparent. During the paradigm cell filling process, the model consistently amplifies the dominance of exponents that outnumber others, setting up a `rich-get-richer', or `preferential attraction' dynamic in which the frequency of the most frequent items progressively rises further, even when doing so causes the loss of other variants. Because this feedback loop proceeds unchecked (since the model contains no disruptive process which would upset it) the model is guaranteed to culminate in a radically homogenised state, with one inflection class for all lexemes. Moreover, our careful re-examination of the dynamic relationships unfolding between cells highlights the limits of conditional entropy as a measurement of predictability, in systems that are undergoing change.

\subsection{A variant model for metamorphomes}\label{existing-metamorphomes}

The mechanisms of Ackerman \& Malouf’s model are also adapted by \citet{Esher2015a,Esher2015b,Esher2017} to represent change affecting metamorphomes (groups of paradigm cells identified by similarity of exponence) as opposed to inflection classes (groups of lexemes identified by similarity of exponence). The input is once again a set of 100 lexemes, each with 8 cells. Each of the 800 total cells is randomly populated with one of the three abstract allomorph indices \citep[see e.g. ][]{Stump2015}, labeled $x$, $y$ and $z$. Thus, the intended interpretation of a pair of lexemes which both contain the index $x$ in cells $A$ and $B$, and the index $y$ in all other cells, is that each lexeme has two allomorphs, one shared by cells $A$ and $B$, and one found in all other cells; the phonological content of the allomorph indexed by $x$ may differ between lexemes, but the distributional pattern is constant. At the outset, each of the 100 lexemes typically presents a unique distributional pattern, and the majority of lexemes instantiate 3 different indices.

\begin{figure}[ht]
\caption[Paradigm cell filling mechanism]{Paradigm cell filling mechanism. Rows are lexemes, columns are cells; the focal cell of the focal lexeme, marked `?', is to be filled. (a) Select a pivot cell and examine its contents in all lexemes. (b) Compare pivot and focal cells. (c) Distinguish identity versus contrast lexemes (grey versus black). (d) Sample a lexeme. If identity: in the focal lexeme, copy the focal exponent from the pivot; if contrast: copy the focal exponent at random within the focal lexeme.}
\centering
\includegraphics[width=10cm]{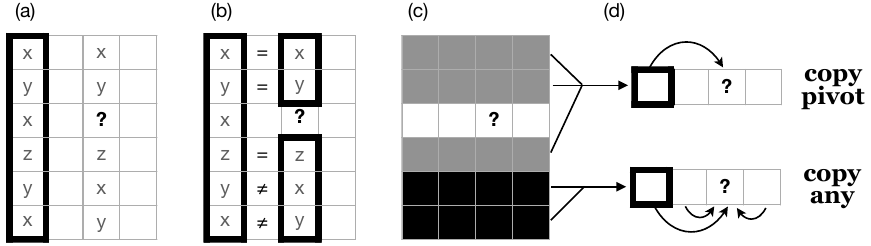}
\label{pcfp-Esher}
\end{figure}

As before, each cycle of the model predicts the `held out' focal exponent by examining evidence from other lexemes (Figure~\ref{pcfp-Esher} stage (a)). In this case, the model does not directly compare indices for the pivot cell across lexemes, but instead probes the \textit{relationship} between the pivot and focal cells (Figure~\ref{pcfp-Esher} stage (b)). Lexemes are classified into one of two groups: an `identity’ group in which the pivot and focal cells share their index, and a `contrast’ group in which the pivot and focal cells have differing indices (Figure~\ref{pcfp-Esher} stage (c)). Two strategies are then available for predicting the focal exponent (Figure~\ref{pcfp-Esher} stage (d)). One, motivated by the `identity’ group, ensures that the focal lexeme also has identical indices in the pivot and focal cells, by copying the pivot's index into the focal cell: this option simulates speakers generalising an abstract distributional pattern (`pivot cell and focal cell share exponents’) onto the focal lexeme. The second strategy, motivated by the `contrast’ group, is intended to simulate a situation in which speakers’ only intuition about the relationship between the pivot and focal cells is that their exponents are not reliably the same: in such a situation, there is no positive basis for informed choice of an index, and the model therefore makes a random choice from among all of the indices present in the other (i.e. non-focal) cells of the focal lexeme. The choice between the two strategies is decided probabilistically, where the probability of each is calculated as the lexical type frequency of the `identity’ and `contrast’ groups respectively.\footnote{In practice, this mechanism is exactly equivalent to picking a single evidence lexeme $l_E$ at random and choosing a strategy according to whether $l_E$ presents `identity’ or `contrast’.} In this model, there is no `tidying-up' which deletes duplicate lexical items; the model simply proceeds to its next cycle, and continues to iterate until 25 cycles have passed without change. 100 simulation runs are conducted for each of 10 initial lexicons.

\begin{figure}[ht]
\caption{Replication of \citeauthor{Esher2015a}'s (\citeyear{Esher2015a,Esher2015b}) model. Evolution of 100 initial lexemes with 8 cells and 5 available allomorph indices. 100 simulation runs of 30,000 cycles. `Morphomic zones' are sets of cells which, in every lexeme, share their index.}
\centering
\includegraphics[width=11cm]{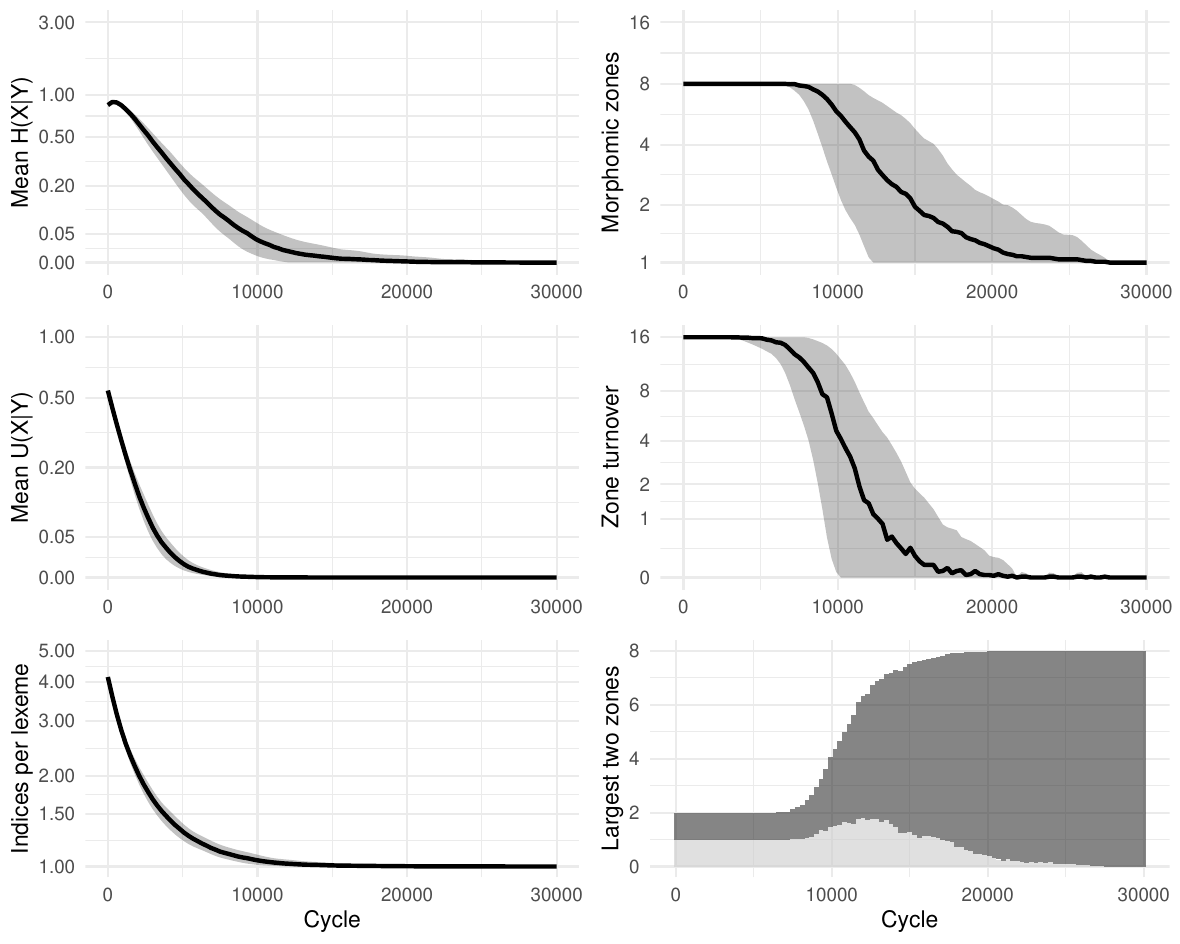}
\label{plot-Esher}
\end{figure}

\begin{figure}[ht]
\caption{Eight snapshots evenly spaced between cycle 0 (leftmost) and cycle 30,000 (rightmost) from one simulation of \citeauthor{Esher2015a}'s (\citeyear{Esher2015a,Esher2015b}) model. Each snapshot shows 100 lexemes in rows, 8 cells in columns. Distinct allomorph indices in each lexeme are indicated by shading.}
\centering
\includegraphics[width=12cm]{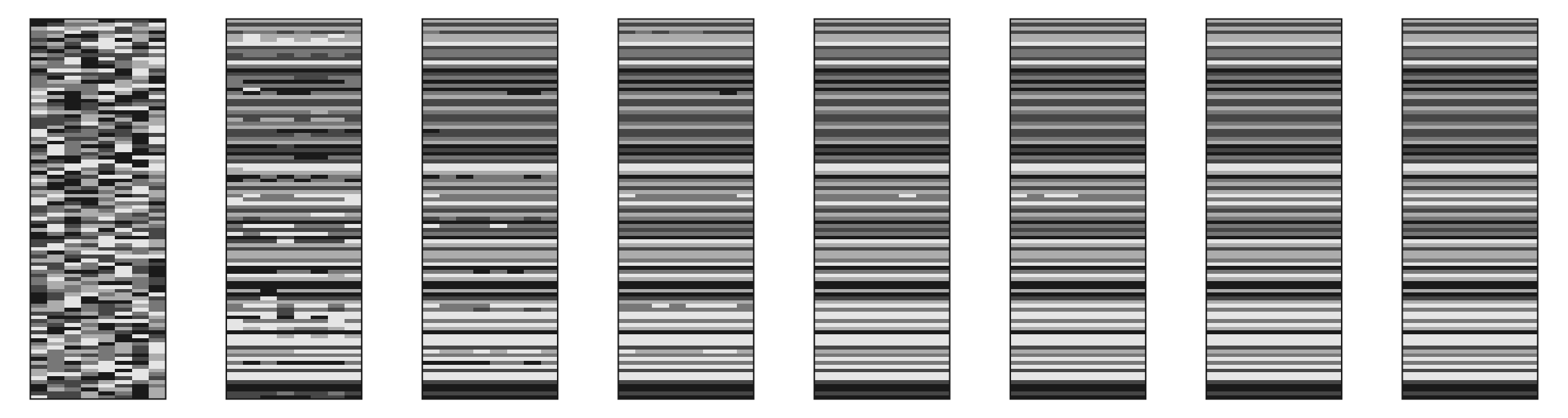}
\label{snapshots-Esher}
\end{figure}

Esher’s model, like Ackerman \& Malouf’s model, consistently reports a reduction in inflectional diversity (in the `final’ state, most lexemes present the same index in all cells, or a default/exception pattern in which one cell has a different index to the 7 others), and for the same reason of radical homogenisation. It is significant that, for lexemes exhibiting the default/exception pattern in `final’ states of Esher’s model, the exceptional cell is a different one in each lexeme: thus the model demonstrably does not lead to consistency of patterning across lexemes, in contrast to what is observed for metamorphomes in natural language. We constructed a replica of this model, tracking distributional patterning, mean conditional entropy and Theil’s U, and allowed the replica to run indefinitely. Results appear in Figure~\ref{plot-Esher} and snapshots from one of the simulations are shown in Figure~\ref{snapshots-Esher}. The results confirmed that the `default/exception’ pattern was simply a transitional stage preceding total uniformity of all cells in all lexemes; mean conditional entropy decreased, but mean Theil’s U also fell to zero consistently, indicating that mutual predictability between cells reduced over time. The dynamic in Esher’s model, just as in Ackerman \& Malouf’s, promotes a decline in variation leading to radical homogenisation, as opposed as promoting structured, mutual predictability between cells.%

\subsection{Discussion of the existing models}\label{existing-discussion}

The models developed by \citet{AckermanMalouf2015} and \citet{Esher2015a,Esher2015b,Esher2017} illustrate how a simple iterative simulation can be used to investigate inflectional change. Due to their similar architecture, the models present similar conceptual issues: the only changes allowed are those in which lexemes or cells become more like others, and thus the overall dynamic is one of preferential attraction towards exponents that are already more frequent than their competitors. To the extent that these models exhibit self-organisation, it is only of a radically homogenising kind resulting in complete uniformity: all lexemes inescapably converge on a single inflection class, and all cells on a single metamorphome. Thus, neither model evolves the structured predictability and stable diversity which is amply observed in natural language inflectional systems.

In both models, an additional tension exists between the nature of the paradigm cell filling task, which assumes speaker uncertainty about one inflectional form, and the predictive strategies exploited to complete the task, which assume exhaustive knowledge of all other inflectional forms in the system. \cites{AckermanMalouf2015} model selects lexemes of interest by scanning all lexemes other than the focal lexeme; the model thus has full knowledge of exponents associated with the pivot cell in every lexeme, and of exponents associated with the focal cell in every lexeme bar the focal lexeme. At the same time however, the model myopically examines cells only two at a time, making no reference to cells beyond the focus and the pivot. Esher’s model presents an additional discordant note: reference is made to the non-focal cells of the focal lexeme in the `contrast’ strategy, but not in any other conditions. We return to these issues in Sections \ref{family} and \ref{NE-model}, enriching the existing models with additional parameters motivated by observations of natural language.

\FloatBarrier
\section{A family of enhanced models}\label{family}

Building on our replicas of the earlier models, and in order to address the observed limitations of these, we now present a series of related models incorporating additional parameters. Our innovations are focused on two specific aspects of the models: firstly, the principles by which the paradigm cell filling task is accomplished given a set of evidence, and secondly, the selection and weighting of evidence which feeds into it. We implement these extensions as independently modulable parameters to facilitate the controlled observation of their effects. The resulting family of models is implemented in a package \texttt{paradigmEvo2023} \citep{paradigmEvo2023} in R \citep{r2024}. To aid comparability, in our model runs below we retain the same initial input, consisting of a lexicon of 100 lexemes and 8 cells in which paradigms are populated with randomly distributed exponents. Likewise, in each cycle, the model must predict the focal exponent based on evidence from other lexemes, and the model’s prediction is integrated into the lexicon input to the next cycle. Internally, the models operate at the same level of abstraction and idealisation as before: exponents are represented as indices rather than phonological forms; each cycle represents a general change in the system as a whole; there is no representation of individual speakers; and the models lack disruptive processes. For simplicity of exposition, we concentrate here on rhizomorphomes (i.e., inflection classes), although our study systematically examined parallel equivalent models for both rhizomorphomes and metamorphomes. Results of the metamorphome simulations appear in Appendix A. Code for our analyses can be found in Supplementary Materials available on zenodo, \textsc{doi}~\href{10.5281/zenodo.13934423}{https://doi.org/10.5281/zenodo.13934423}.

\subsection{Frequency and speaker knowledge}\label{family-frequency}

The approaches of existing models both overestimate and underestimate the range of data to which speakers have access when predicting inflectional forms within natural language systems. We begin by considering the amount of evidence which speakers might use during paradigm cell filling, and where in the lexicon this evidence is drawn from.

\begin{figure}[ht!]
\caption{Sampling 50\% of evidence lexemes. 100 simulations of 10,000 cycles.}
\centering
\includegraphics[width=10.5cm]{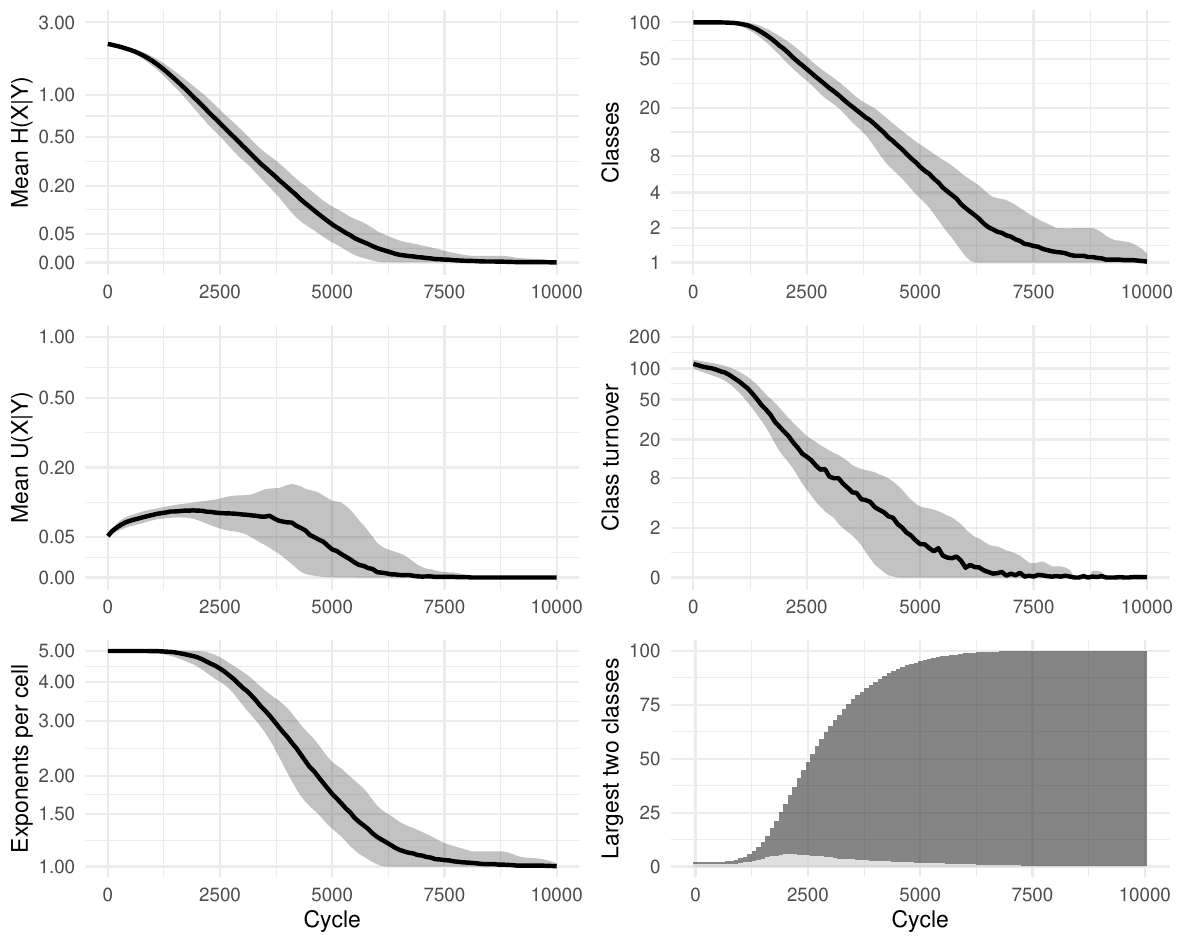}
\label{plot-sample-ev-50pc}
\end{figure}

\begin{figure}[ht!]
\caption{Sampling 20\% of evidence lexemes. 100 simulations of 10,000 cycles.}
\centering
\includegraphics[width=10.5cm]{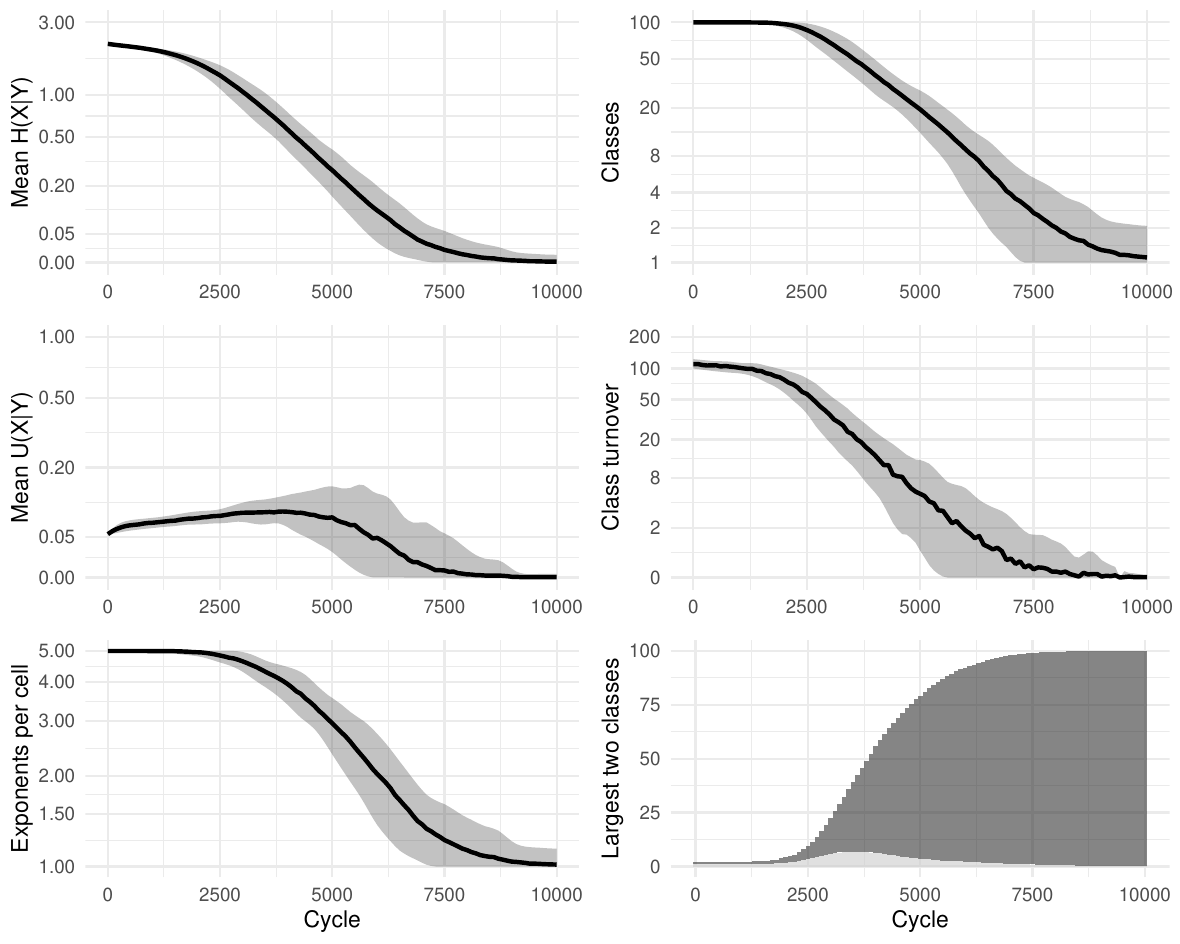}
\label{plot-sample-ev-20pc}
\end{figure}

In our paradigm cell filling model, inductive inference of an unknown focal exponent is based on evidence gathered from  knowledge of exponents in other lexemes. Recent research indicates that, when facing inductive problems, humans do not evaluate all memorised evidence, but instead draw smaller samples and reason on the basis of these \citep{vul2014one,shenhav2017toward}. We incorporate this insight into our models by allowing evidence lexemes to be sampled. Running each simulation 100 times, we find only minimal differences in outcome between attending to 100\% (cf Figure~\ref{plot-AM-lexemewise} above) and 50\% (Figure~\ref{plot-sample-ev-50pc}) or 20\% (Figure~\ref{plot-sample-ev-20pc}) of the evidence lexemes. The sole effect of sampling fewer lexemes is that the overall evolutionary trajectory takes slightly longer: for instance, the largest inflection class takes around 2,300 cycles on average to dominate half the lexicon with a 100\% sample, around 2,700 cycles with 50\%, and around 4,100 with 20\%.

\begin{figure}[ht!]
\caption{Sampling 20\% of evidence lexemes, using 2 pivots. 100 simulations of 10,000 cycles.}
\centering
\includegraphics[width=10.5cm]{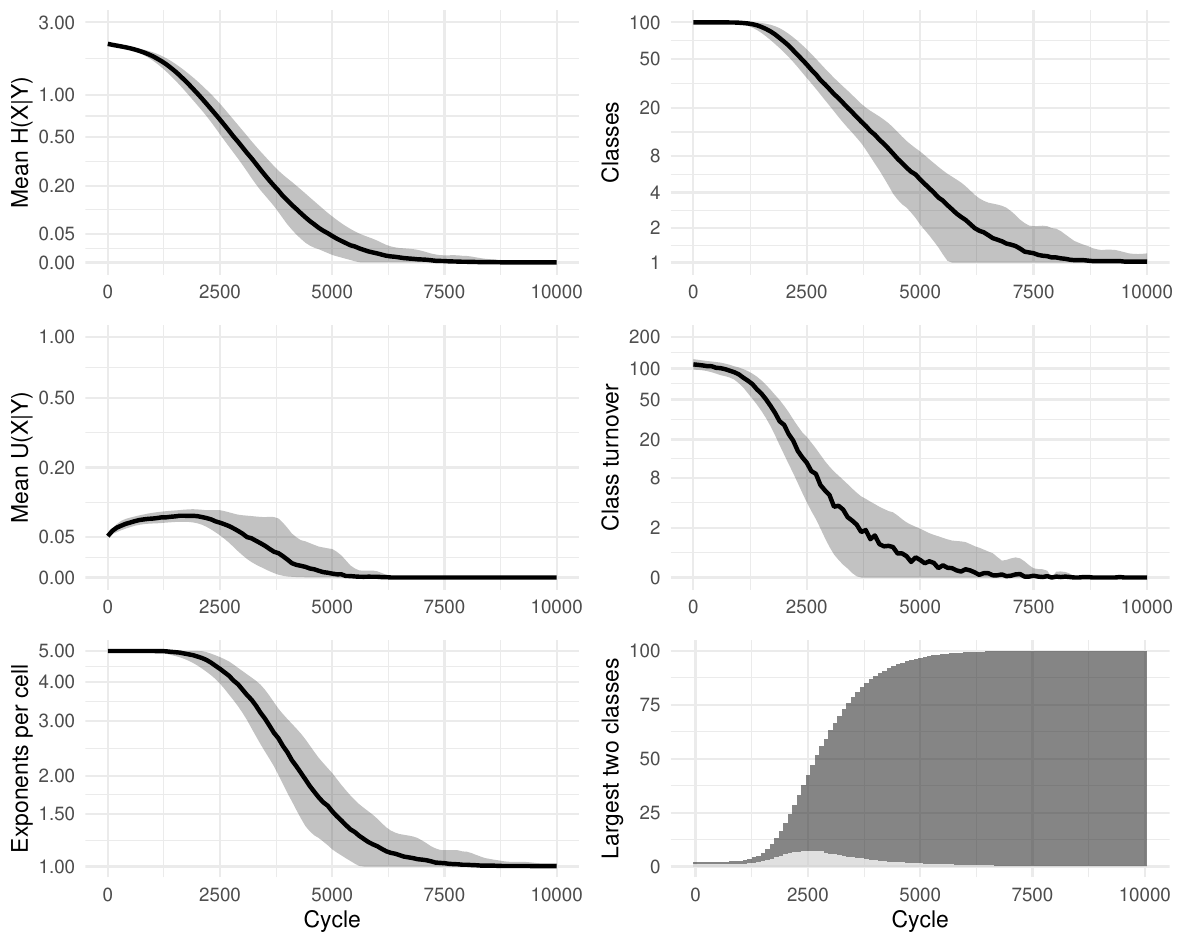}
\label{plot-2pivots-20pc}
\end{figure}

\begin{figure}[ht!]
\caption{Sampling 20\% of evidence lexemes, using 4 pivots. 100 simulations of 10,000 cycles.}
\centering
\includegraphics[width=10.5cm]{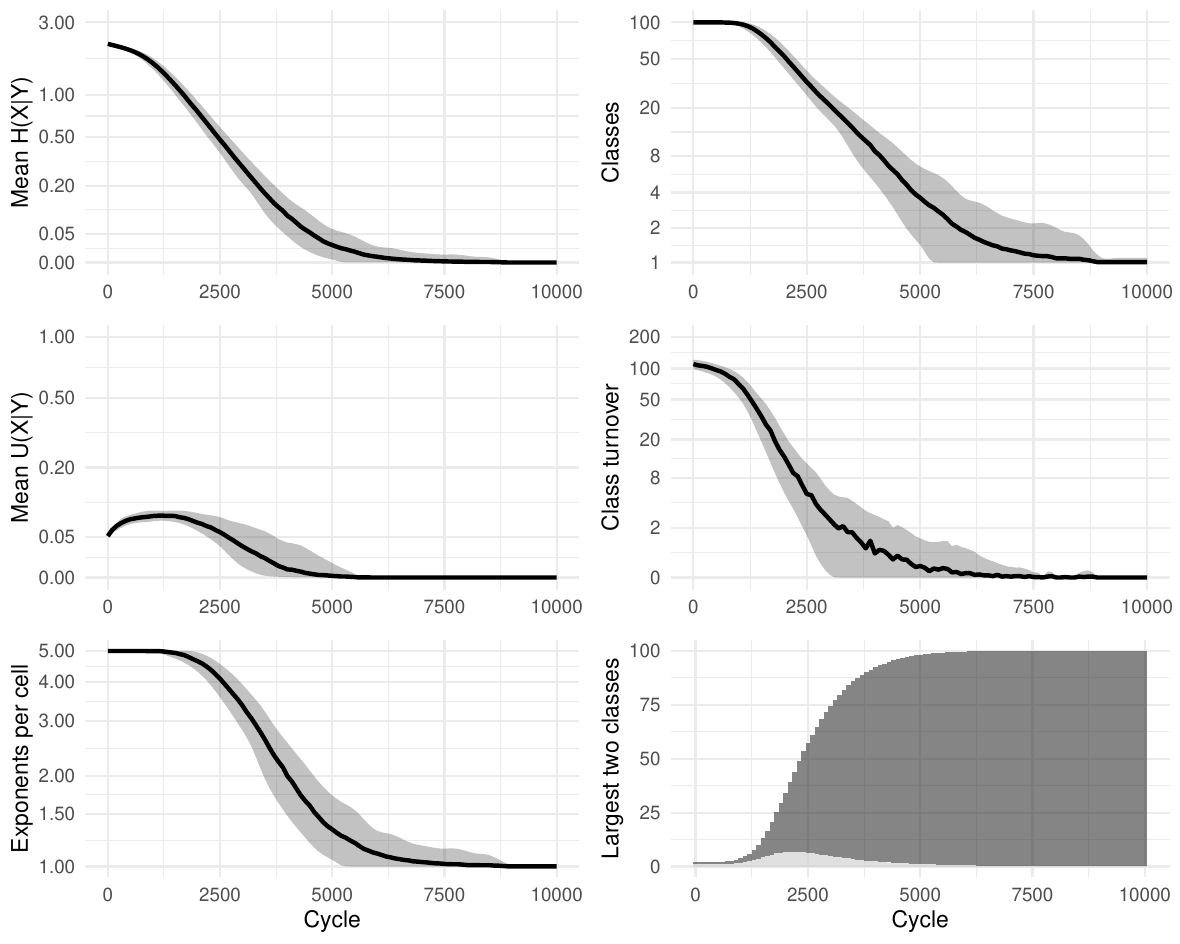}
\label{plot-4pivots-20pc}
\end{figure}

As we have considered it until now, the cell filling process examines only one pivot cell in order to make a selection of potential evidence lexemes. However, it is unlikely that speakers confine themselves to relying on a single pair of cells during prediction (see \citealt{BonamiBeniamine2016} for the increased reliability of predictions based on multiple cells; also \citealt{StumpFinkel2013} for dynamic principal parts). Accordingly, we allow our enhanced models to use multiple pivots. For each individual pivot, evidence is gathered as in Figure~\ref{pcfp-positive-only}, stages (a)–(c), and then the independent evidence from all pivots is summed in order to select the focal exponent. In Figures \ref{plot-2pivots-20pc} and \ref{plot-4pivots-20pc} we show results of using 2 and 4 pivots while sampling 20\% of the available evidence lexemes; cf Figure~\ref{plot-sample-ev-20pc} above, using 1 pivot. As with the sampling of evidence lexemes, the principle difference is that as the model considers more information, the evolutionary trajectory proceeds more rapidly. For instance, the largest inflection class takes around 4,100 cycles to dominate half the lexicon using one pivot, around 2,900 cycles with two pivots, and around 2,500 with four.

In real speech, the frequencies at which any given lexemes or paradigm cells occur can vary starkly, and is approximately Zipfian \citep{BlevinsMilinEtAl2016}. Consequently, speakers’ likelihood of having heard a form, and thus being able to use it as evidence during the cell filling task, will follow a probability distribution that is non-uniform. Moreover, \citet{simswilliams2022} shows that the likelihood of a cell undergoing analogical change (in our model: being chosen as the focal cell) is negatively correlated with its token frequency. To reflect these facts, we allow a Zipfian distribution to be placed over lexemes and over cells: more frequent items are more favoured as evidence to solve the PCFP (on the assumption that speakers can more readily access them), while less frequent items are more likely to be the focus of the inference (on the assumption that they are less likely to be known, thus more likely to require recourse to analogical reasoning during production). Notionally, Zipfian skewing could take the form of weighting evidence, or of sampling it. For instance, we might attend to all evidence lexemes (or cells) but then weight them in a Zipfian manner; or we might sample a subset of all lexemes (or cells) with Zipfian probabilities but thereafter weight them evenly. In our model we implemented both options, though for reasons of space here we discuss only the sampling option.\footnote{Mathematically speaking, the two options would have essentially identical long-term average behaviour, however the sampling option will exhibit more random variation in the short term, and this could lead to interesting qualitative differences of behaviour. Since evaluating the precise impact of alternative implementations of Zipfian skewing would take us beyond the scope of this paper, we leave the question for future research.}

\begin{figure}[ht!]
\caption{Sampling 20\% of evidence lexemes, using 2 pivots, sampling cells according to a Zipfian distribution. 100 simulations of 50,000 cycles.}
\centering
\includegraphics[width=10.5cm]{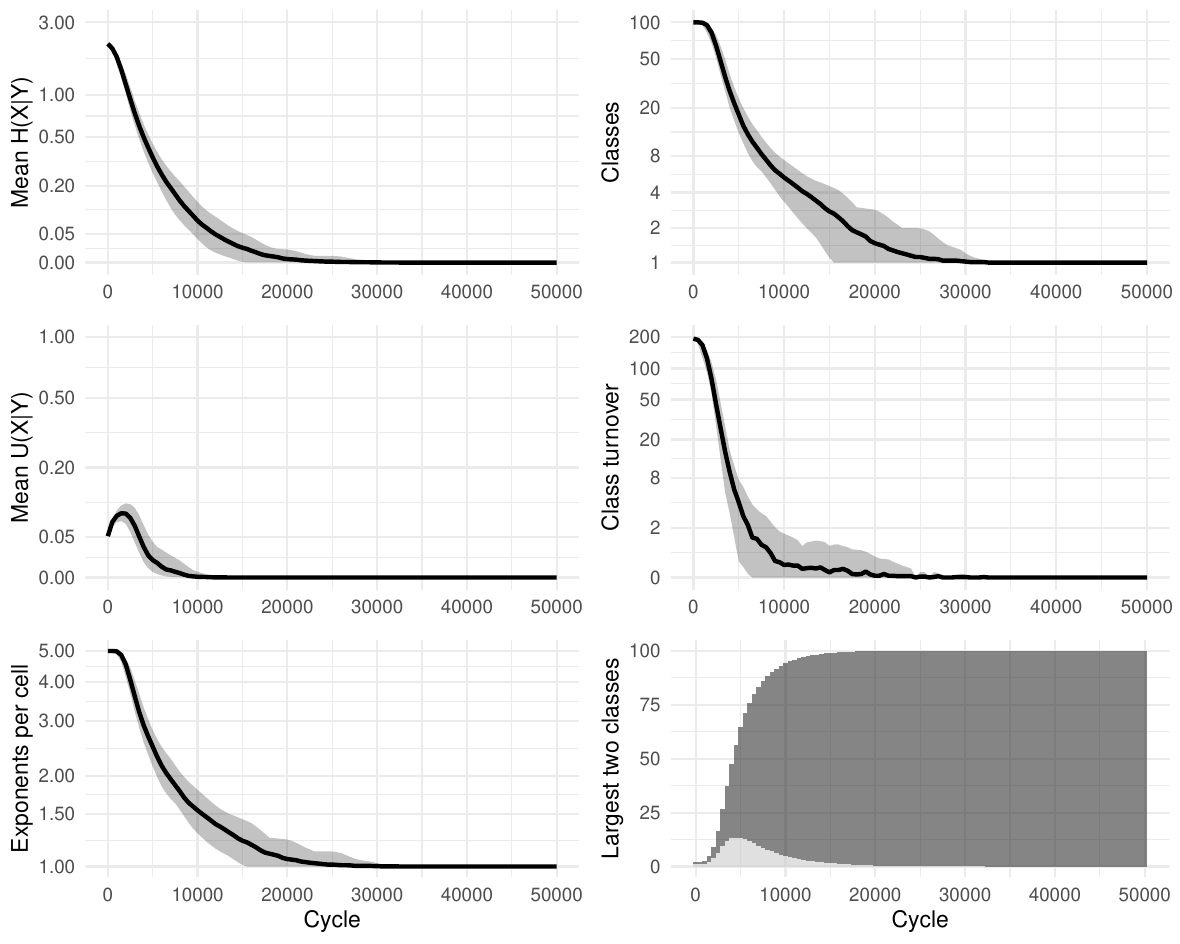}
\label{plot-zipfian-cells-20pc}
\end{figure}

Zipfian sampling has a significant impact. Figure~\ref{plot-zipfian-cells-20pc} shows the results of sampling cells---both foci and pivots---in a Zipfian manner, using 2 pivots and sampling 20\% of evidence lexemes (compare Figure~\ref{plot-2pivots-20pc} which also uses 2 pivots and a 20\% sample, but samples cells uniformly). Most significantly, the shift to Zipfian sampling entails that some cells are only very rarely selected as the focal cell, and consequently are very slow to change. The overall effect is to slow down the evolutionary trajectory of the system, particularly in its final approach to uniformity. The system still quickly reduces to only a few classes: from the initial 100 to an average of 5 classes by 10,000 cycles and to an average of just 2 classes by 20,000 cycles, with the largest inflection class covering 50\% of lexemes by around 5,100 cycles. However, owing to rarity of changes to the least-sampled cell, the final steps to uniformity are drawn out until as many as 30,000 cycles.

\begin{figure}[ht!]
\caption{Sampling 20\% of evidence lexemes, using 2 pivots, sampling lexemes according to a Zipfian distribution. 100 simulations of 100,000 cycles.}
\centering
\includegraphics[width=10.5cm]{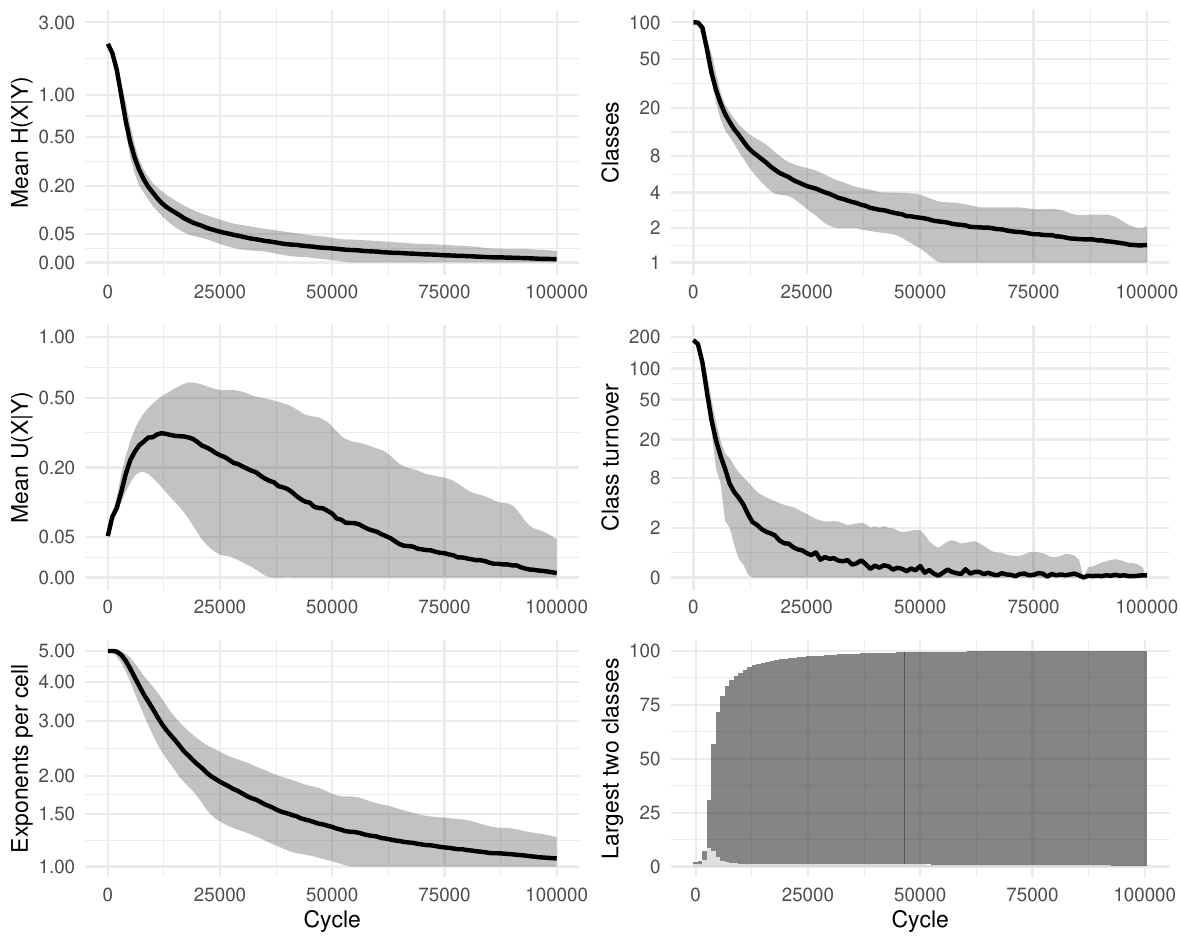}
\label{plot-zipfian-lexemes-20pc}
\end{figure}

Figure~\ref{plot-zipfian-lexemes-20pc} shows the results of sampling lexemes---both focal and evidence lexemes---in a Zipfian manner, and using 20\% of evidence lexemes. The final stages of the evolutionary process are significantly slowed down, since some of the 100 lexemes are only rarely chosen as the focal lexeme. In Figure~\ref{plot-zipfian-lexemes-20pc} simulations run for 100,000 cycles, at which point around 40\% of the runs still retain two inflection classes. On the other hand, the largest inflection class already covers 50\% of lexemes after just 5,000 cycles, emphasising the fact that primarily it is only the very tail end of the evolutionary process which is extended through the addition of Zipfian sampling.

Consistently, we find that incorporating more realistic frequency distributions has the effect of slowing the models’ progression to uniformity, but that the progression itself remains inevitable. Moreover, stable inflectional structure still does not emerge in any of our enhanced models, whether for rhizomorphomes (shown here) or for metamorphomes (see Figures A3 and A4 in Appendix A). For instance, there is no case in which we see emergence of three or four sizeable, distinctive inflection classes. Rather, in all cases, both mean conditional entropy and mean Theil’s U decrease as the models run, indicating that the reduction in mean conditional entropy is due to loss of variation rather than to emergence of structured variation and mutual predictiveness between cells. That is, just as in the original iterative models of Section \ref{existing}, we find little conformity between the behaviour of the enhanced models and the autonomous morphology which they are intended to simulate. This finding is striking given that the paradigm cell filling task in the enhanced models is considerably more realistic: despite incorporating multiple insights from the empirical observation of natural language, the enhanced models still do not generate the stable persistence of structured variation observed in real-life inflectional systems.

\subsection{Associative evidence and the pure-attraction dynamic}\label{family-positive}

The enhanced model results show that even with more realistic parameters for sampling, the systems are still only capable of moving from initial, unstructured randomness to a stable state of complete uniformity. In all cases, a single inflection class rises rapidly to dominance, and as it extends its grip to the final corners of the lexicon, variation across lexemes is eliminated for one paradigm cell after another. At this point, Theil’s U ceases to climb and begins falling, indicating that the evolutionary dynamic has shifted from one of increasing predictability between cells to one in which steadily decreasing diversity means there is little left to predict. Although Zipfian sampling will prolong the final act in which the last few lexemes succumb to the dominant class, it does not promote any extended, earlier period in which multiple inflection classes thrive together. In this sense, we do not find any of the models examined so far to correspond well to the persistence of autonomous morphology observed in natural language.

As a step towards deeper understanding, let us clarify the fundamental dynamic which causes all the systems considered above to follow the same inexorable trajectory towards uniformity. The rapid rise to dominance of a single inflection class is the automatic outcome of a cycle-by-cycle dynamic of preferential attraction. This dynamic is itself guaranteed by the nature of the paradigm cell filling process as we have considered it so far: at every change, some lexeme becomes more similar to another. The inevitability of increasing similarity at every step follows from the underlying analogical process which solely seeks out \textsc{associative evidence}, that is, evidence for similarity between lexemes, at step (b) in Figure~\ref{pcfp-positive-only}. Having found some similarity, the process then adds more, creating an attraction-only dynamic between lexemes, which leads eventually to their complete convergence. No system evolving under these conditions will create and settle into a stable diversity of morphomes; it will always tend towards the reduction of diversity, on a path to complete uniformity.

Such is the power of this dynamic that even the introduction of disruptive events (such as sound changes) would not counteract it with lasting effect. The reason is as follows. A disruptive event has the potential to cause an increase in the number of inflection classes present in the lexicon at the point where it applies. However, in the context of our models, this is equivalent to returning the inflectional system to an earlier state in its overall path towards uniformity.\footnote{A caveat is that we assume that the sound change does not result in a total partitioning of the lexicon in the sense we introduced in Section~\ref{existing-variations}. A partitioning would require some set of lexemes to acquire exponents in \textit{every} paradigm cell, none of which are found outside of that set of lexemes. We suspect that this would be highly unlikely in inflectional systems with a large number of paradigm cells, though it would be more plausible in systems with very few cells. Importantly in this regard, we note that empirically speaking, morphomic structure is not a phenomenon restricted to systems with only few paradigm cells, but rather is regularly found in systems with very large paradigms. Thus, even if partitioning due to sound changes could account for some cases of morphomic stability, it will not account for all.}
As soon as the analogical process resumes, similarity will increase again, and the temporary spike in diversity induced by disruption will be quickly ironed out without trace. While frequent disruptions could perpetually introduce unstructured variation, this would not result in stable morphomic persistence, such as is frequently observed in natural languages. In sum, we observe that it is impossible for a system governed by a pure-attraction dynamic, even one punctuated by disruptions, to evolve new, resilient morphomic diversity.

\FloatBarrier
\section{Dissociative evidence and an attraction-repulsion dynamic}\label{NE-model}

In order to discover a path to stable diversity of the type observed in natural language, we must re-examine assumptions about analogical reasoning.

The models discussed so far all attend to associative evidence, that is, evidence which is based on similarities among lexemes and which suggests what the focal exponent \textit{should} be. However, in doing so, the paradigm cell filling process consistently discounts other evidence which is potentially available within the system. Namely, a rational agent also has access to \textsc{dissociative evidence}, i.e. evidence based on dissimilarities among lexemes which suggests what the focal exponent \textit{should not} be. There is no \textit{a priori} reason for such dissociative evidence to be ignored.

We begin by showing how dissociative evidence arises during analogical reasoning in Section~\ref{why-dissociative}. In Section~\ref{PCF-revised} we integrate dissociative evidence into the paradigm cell filling process and examine its effects on the evolution of the inflectional system.

\subsection{How dissociative evidence arises}\label{why-dissociative}

In the models considered in this paper, the smallest units of decision making are simple, four-part analogies. For instance, the model of \citet{AckermanMalouf2015} carries out comparisons of a focal cell and a pivot cell in a focal lexeme and an evidence lexeme, which could be set out as in (1), where the exponent in the focal cell of the focal lexeme is to be inferred. The model may carry out many of these atomic comparisons and then sum up their individual answers, but the most basic decisions being made are simple, four-part analogies.

\begin{table}[H]
\begin{tabular}{llclc}
(1) &                 & Pivot cell &    & Focal cell \\
    & Evidence lexeme & A          & :: & B          \\
    & Focal lexeme    & C          & :: & \textbf{?}       
\end{tabular}
\end{table}

Recent research in experimental and computational cognitive science has argued that human inferential reasoning is not a one-size-fits-all affair, but rather is sensitive to its \textit{inductive context} which may differ according to the information available and the nature of the task \citep{kemp2009structured,kemp2014taxonomy}. In \citet{AckermanMalouf2015}, the four-part inferential calculus is as follows: if the exponents in the two pivot cells are the same, then exponents in the two focal cells should be the same. If the exponents in the two pivot cells are different, then no prediction is made. However, here we show that this reasoning in fact fails to make full use of the information available to a four-part reasoner. Namely, when the exponents in the two pivot cells are different, then the most rational decision is that the exponents in the two focal cells should also differ. To show why this is true, we first consider a specific scenario and thereafter shift to the general case. By doing so, our aim is to provide an initial, intuitive sense of the logic first, and thereafter move to a conclusive argument.

The four-part reasoner has extremely limited information to work with. It knows nothing more than the exponents of the two pivot cells, and the exponent of one focal cell. However, from this basis, more can be inferred. To see why this is so, we consider an inflectional system in which our attention is narrowed to just two cells, which we refer to as cell 1 and cell 2, and just two lexemes: the evidence lexeme for which we know the contents of cells 1 and 2, and the focal lexeme for which we know only the contents of cell 1.

We begin with a specific case. We will generalise away from the specifics later, but for now, suppose that cell 1 has three possible exponents, denoted $a$,  $b$ and $c$, and that cell 2 also has three possible exponents, denoted $x$, $y$ and $z$. 

Reasoning about the unknown exponent (in cell 2 of the focal lexeme) requires the reasoner to keep track of which \textit{possible combinations} a lexeme can have, of an exponent in cells 1 and an exponent in cell 2. To keep track of these two-cell combinations, we can use a simple, two-dimensional contingency table as in (2). Within the table, we place a `1' to indicate that the reasoner \textit{knows} that a certain combination is possible in the language. For instance, if the evidence lexeme contained the exponent combination $\langle b,z \rangle$ in cells 1 and 2, then the reasoner could begin to fill in the contingency table as in (3). When the reasoner is unsure whether a combination is permitted, we will use fractional numbers to indicate the probability that the reasoner is able to assign to the possibility that it is permitted.

\begin{table}[H]
\begin{tabular}{cc|c|c|c|p{0.6cm}cc|c|c|c|}
\cline{3-5} \cline{9-11}
(2)                   &     & $x$ & $y$ & $z$ &  & (3)                   &     & $x$ & $y$ & $z$ \\ \cline{2-5} \cline{8-11} 
\multicolumn{1}{c|}{} & $a$ &     &     &     &  & \multicolumn{1}{c|}{} & $a$ &     &     &     \\ \cline{2-5} \cline{8-11} 
\multicolumn{1}{c|}{} & $b$ &     &     &     &  & \multicolumn{1}{c|}{} & $b$ &     &     & 1   \\ \cline{2-5} \cline{8-11} 
\multicolumn{1}{c|}{} & $c$ &     &     &     &  & \multicolumn{1}{c|}{} & $c$ &     &     &     \\ \cline{2-5} \cline{8-11} 
\end{tabular}
\end{table}

To elaborate our specific case further, suppose that the evidence lexeme contains the combination $\langle a,x \rangle$, so the reasoner begins filling in the contingency table as in (4). Moreover, suppose that the inflectional system permits only three possible combinations of an exponent in cell 1 and an exponent in cell 2. What can a reasoner infer from this information? We know that the inflectional system uses exponents $a$, $b$ and $c$ in cell 1, so there will have to be at least one combination that contains $a$ in cell 1, at least one with $b$ and at least one with $c$; likewise, the inflectional system uses exponents $x$, $y$ and $z$ in cell 2, so there will have to be at least one combination that contains $x$ in cell 2, at least one with $y$ and at least one with $z$. Under these tight constraints, the system can correspond to only two \textit{possible} alternatives: either (5) or (6). 

\begin{table}[H]
\begin{tabular}{cc|c|c|c|p{0.6cm}cc|c|c|c|p{0.6cm}cc|c|c|c|}
\cline{3-5} \cline{9-11} \cline{15-17}
(4)                   &     & $x$ & $y$ & $z$ &  & (5)                   &     & $x$ & $y$ & $z$ &  & (6)                   &     & $x$ & $y$ & $z$ \\ \cline{2-5} \cline{8-11} \cline{14-17} 
\multicolumn{1}{c|}{} & $a$ & 1   &     &     &  & \multicolumn{1}{c|}{} & $a$ & 1   &     &     &  & \multicolumn{1}{c|}{} & $a$ & 1   &     &     \\ \cline{2-5} \cline{8-11} \cline{14-17} 
\multicolumn{1}{c|}{} & $b$ &     &     &     &  & \multicolumn{1}{c|}{} & $b$ &     & 1   &     &  & \multicolumn{1}{c|}{} & $b$ &     &     & 1   \\ \cline{2-5} \cline{8-11} \cline{14-17} 
\multicolumn{1}{c|}{} & $c$ &     &     &     &  & \multicolumn{1}{c|}{} & $c$ &     &     & 1   &  & \multicolumn{1}{c|}{} & $c$ &     & 1   &     \\ \cline{2-5} \cline{8-11} \cline{14-17} 
\end{tabular}
\end{table}

Although the reasoner does not have enough information to know with certainty which of alternatives (5) or (6) is true, it can infer that one of them must be true. Since it has no additional information for choosing between the two alternatives, the most neutral stance is to accord an equal probability to each. As a consequence, the reasoner can now assign probabilities to all cells of the contingency table as shown in (7).

\begin{table}[H]
\begin{tabular}{cc|c|c|c|}
\cline{3-5}
(7)                   &     & $x$ & $y$ & $z$ \\ \cline{2-5}
\multicolumn{1}{c|}{} & $a$ & 1   & 0   & 0   \\ \cline{2-5}
\multicolumn{1}{c|}{} & $b$ & 0   & 0.5 & 0.5 \\ \cline{2-5}
\multicolumn{1}{c|}{} & $c$ & 0   & 0.5 & 0.5 \\ \cline{2-5}
\end{tabular}
\end{table}

Consider now what table (7) implies. It states that in this language, if a lexeme has exponent $a$ in cell 1, it will necessarily have $x$ in cell 2. If it has something other than $a$ in cell 1, it will necessarily have something other than $x$ in cell 2 (with the two alternatives, $y$ and $z$, being equally likely). We reached this conclusion by reasoning from an initial observation of the combination $\langle a,x \rangle$ in the evidence lexeme. Suppose instead that the evidence lexeme contained $\langle b,z \rangle$. The conclusion then would be that if a lexeme has $b$ in cell 1 it has $z$ in cell 2, and if it has other than $b$ in cell 1 it has other than $z$ in cell 2. In the general case, the conclusion will always be: if a lexeme is the same in cell 1 as the evidence lexeme, it is the same in cell 2 as the evidence lexeme (this is associative evidence); and if a lexeme is different in cell 1 from the evidence lexeme then it is different in cell 2 from the evidence lexeme (this is dissociative evidence). Taken together, the associative and dissociative evidence sum up to a simple heuristic: same in cell 1 implies same in cell 2; different in cell 1 implies different in cell 2.

Next let us relax some of the specifics of our case study. Suppose that instead of only three possible combinations of exponents in cells 1 and 2, the language permits four. We observe combination $\langle a,x \rangle$ in the evidence lexeme. The question for the reasoner is, `what specific, alternative possibilities is this now compatible with?' There are 20 possibilities, shown in Figure \ref{twenty-alternatives}. If the reasoner is neutral and treats each as equally probable, then the total probabilities are as shown in (8).

\begin{figure}[ht!]
\caption{All 20 alternative possibilities of allowing four combinations of one of $\{a, b, c\}$ in cell 1 and one of $\{x,y,z\}$ in cell two, given that $\langle a,x \rangle$ is among them.}
\footnotesize % Make the font small
\begin{tabular}{c|c|c|c|p{0.5cm}c|c|c|c|p{0.5cm}c|c|c|c|p{0.5cm}c|c|c|c|}
\cline{2-4} \cline{7-9} \cline{12-14} \cline{17-19}
                          & $x$ & $y$ & $z$ &                       &     & $x$ & $y$ & $z$ &                       &     & $x$ & $y$ & $z$ &                       &     & $x$ & $y$ & $z$ \\ \cline{1-4} \cline{6-9} \cline{11-14} \cline{16-19} 
\multicolumn{1}{|l|}{$a$} & 1   & 1   &     & \multicolumn{1}{c|}{} & $a$ & 1   &     &     & \multicolumn{1}{c|}{} & $a$ & 1   &     & 1   & \multicolumn{1}{c|}{} & $a$ & 1   &     &     \\ \cline{1-4} \cline{6-9} \cline{11-14} \cline{16-19} 
\multicolumn{1}{|l|}{$b$} & 1   &     &     & \multicolumn{1}{c|}{} & $b$ & 1   & 1   &     & \multicolumn{1}{c|}{} & $b$ & 1   &     &     & \multicolumn{1}{c|}{} & $b$ & 1   &     & 1   \\ \cline{1-4} \cline{6-9} \cline{11-14} \cline{16-19} 
\multicolumn{1}{|l|}{$c$} &     &     & 1   & \multicolumn{1}{c|}{} & $c$ &     &     & 1   & \multicolumn{1}{c|}{} & $c$ &     & 1   &     & \multicolumn{1}{c|}{} & $c$ &     & 1   &     \\ \cline{1-4} \cline{6-9} \cline{11-14} \cline{16-19} 
\multicolumn{1}{l}{} \\
\cline{2-4} \cline{7-9} \cline{12-14} \cline{17-19}
                         & $x$ & $y$ & $z$ &                       &     & $x$ & $y$ & $z$ &                       &     & $x$ & $y$ & $z$ &                       &     & $x$ & $y$ & $z$ \\ \cline{1-4} \cline{6-9} \cline{11-14} \cline{16-19} 
\multicolumn{1}{|l|}{$a$} & 1   &     &     & \multicolumn{1}{c|}{} & $a$ & 1   & 1   &     & \multicolumn{1}{c|}{} & $a$ & 1   &     & 1   & \multicolumn{1}{c|}{} & $a$ & 1   &     &     \\ \cline{1-4} \cline{6-9} \cline{11-14} \cline{16-19} 
\multicolumn{1}{|l|}{$b$} & 1   &     &     & \multicolumn{1}{c|}{} & $b$ &     &     & 1   & \multicolumn{1}{c|}{} & $b$ &     & 1   &     & \multicolumn{1}{c|}{} & $b$ &     & 1   & 1   \\ \cline{1-4} \cline{6-9} \cline{11-14} \cline{16-19} 
\multicolumn{1}{|l|}{$c$} &     & 1   & 1   & \multicolumn{1}{c|}{} & $c$ & 1   &     &     & \multicolumn{1}{c|}{} & $c$ & 1   &     &     & \multicolumn{1}{c|}{} & $c$ & 1   &     &     \\ \cline{1-4} \cline{6-9} \cline{11-14} \cline{16-19} 
\multicolumn{1}{l}{} \\
\cline{2-4} \cline{7-9} \cline{12-14} \cline{17-19}
                         & $x$ & $y$ & $z$ &                       &     & $x$ & $y$ & $z$ &                       &     & $x$ & $y$ & $z$ &                       &     & $x$ & $y$ & $z$ \\ \cline{1-4} \cline{6-9} \cline{11-14} \cline{16-19} 
\multicolumn{1}{|l|}{$a$} & 1   &     &     & \multicolumn{1}{c|}{} & $a$ & 1   &     &     & \multicolumn{1}{c|}{} & $a$ & 1   & 1   &     & \multicolumn{1}{c|}{} & $a$ & 1   & 1   &     \\ \cline{1-4} \cline{6-9} \cline{11-14} \cline{16-19} 
\multicolumn{1}{|l|}{$b$} &     & 1   &     & \multicolumn{1}{c|}{} & $b$ &     &     & 1   & \multicolumn{1}{c|}{} & $b$ &     & 1   &     & \multicolumn{1}{c|}{} & $b$ &     &     & 1   \\ \cline{1-4} \cline{6-9} \cline{11-14} \cline{16-19} 
\multicolumn{1}{|l|}{$c$} & 1   &     & 1   & \multicolumn{1}{c|}{} & $c$ & 1   & 1   &     & \multicolumn{1}{c|}{} & $c$ &     &     & 1   & \multicolumn{1}{c|}{} & $c$ &     & 1   &     \\ \cline{1-4} \cline{6-9} \cline{11-14} \cline{16-19} 
\multicolumn{1}{l}{} \\
\cline{2-4} \cline{7-9} \cline{12-14} \cline{17-19}
                         & $x$ & $y$ & $z$ &                       &     & $x$ & $y$ & $z$ &                       &     & $x$ & $y$ & $z$ &                       &     & $x$ & $y$ & $z$ \\ \cline{1-4} \cline{6-9} \cline{11-14} \cline{16-19} 
\multicolumn{1}{|l|}{$a$} & 1   & 1   &     & \multicolumn{1}{c|}{} & $a$ & 1   &     & 1   & \multicolumn{1}{c|}{} & $a$ & 1   &     &     & \multicolumn{1}{c|}{} & $a$ & 1   &     & 1   \\ \cline{1-4} \cline{6-9} \cline{11-14} \cline{16-19} 
\multicolumn{1}{|l|}{$b$} &     &     & 1   & \multicolumn{1}{c|}{} & $b$ &     & 1   &     & \multicolumn{1}{c|}{} & $b$ &     & 1   & 1   & \multicolumn{1}{c|}{} & $b$ &     & 1   &     \\ \cline{1-4} \cline{6-9} \cline{11-14} \cline{16-19} 
\multicolumn{1}{|l|}{$c$} &     &     & 1   & \multicolumn{1}{c|}{} & $c$ &     & 1   &     & \multicolumn{1}{c|}{} & $c$ &     & 1   &     & \multicolumn{1}{c|}{} & $c$ &     & 1   & 1   \\ \cline{1-4} \cline{6-9} \cline{11-14} \cline{16-19} 
\multicolumn{1}{l}{} \\
\cline{2-4} \cline{7-9} \cline{12-14} \cline{17-19}
                         & $x$ & $y$ & $z$ &                       &     & $x$ & $y$ & $z$ &                       &     & $x$ & $y$ & $z$ &                       &     & $x$ & $y$ & $z$ \\ \cline{1-4} \cline{6-9} \cline{11-14} \cline{16-19} 
\multicolumn{1}{|l|}{$a$} & 1   &     & 1   & \multicolumn{1}{c|}{} & $a$ & 1   &     &     & \multicolumn{1}{c|}{} & $a$ & 1   &     & 1   & \multicolumn{1}{c|}{} & $a$ & 1   &     &     \\ \cline{1-4} \cline{6-9} \cline{11-14} \cline{16-19} 
\multicolumn{1}{|l|}{$b$} &     & 1   &     & \multicolumn{1}{c|}{} & $b$ &     & 1   & 1   & \multicolumn{1}{c|}{} & $b$ &     &     & 1   & \multicolumn{1}{c|}{} & $b$ &     &     & 1   \\ \cline{1-4} \cline{6-9} \cline{11-14} \cline{16-19} 
\multicolumn{1}{|l|}{$c$} &     &     & 1   & \multicolumn{1}{c|}{} & $c$ &     &     & 1   & \multicolumn{1}{c|}{} & $c$ &     & 1   &     & \multicolumn{1}{c|}{} & $c$ &     & 1   & 1   \\ \cline{1-4} \cline{6-9} \cline{11-14} \cline{16-19} 
\end{tabular}
\normalsize
\label{twenty-alternatives}
\end{figure}

\begin{table}[H]
\begin{tabular}{cc|c|c|c|}
\cline{3-5}
(8)                   &     & $x$  & $y $ & $z$  \\ \cline{2-5}
\multicolumn{1}{c|}{} & $a$ & 1    & 0.25 & 0.25 \\ \cline{2-5}
\multicolumn{1}{c|}{} & $b$ & 0.25 & 0.5  & 0.5  \\ \cline{2-5}
\multicolumn{1}{c|}{} & $c$ & 0.25 & 0.5  & 0.5  \\ \cline{2-5}
\end{tabular}
\end{table}

Now consider what table (8) implies. It states that in this language, if a lexeme has exponent $a$ in cell 1, then the highest-probability exponent in cell 2 is $x$. If a lexeme has something other than $a$ in cell 1, then the equal highest-probability exponents in cell 2 are those other than $x$. Thus the conclusion is much like before, only probabilistic: same in cell 1 implies that same in cell 2 is the most probable; different in cell 1 implies that different in cell 2 is the most probable. Thus if a reasoner is attempting to maximise its chances of guessing correctly, it will follow the same simple heuristic as before: same implies same; different implies different.

We are now ready to generalise this finding fully. For a system in which cell 1 allows $m$ possible exponents (numbered $1,2,\ldots,m$); cell 2 allows $n$ possible exponents (numbered $1,2,\ldots,n$); the language as a whole permits $k$ possible combinations (where $k$ is less than $m \times n$, i.e., not \textit{every} combination is possible); and the evidence lexeme contains the exponent combination $\langle i,j \rangle$, the probabilities in the contingency table will be as shown in (9), namely: probability of 1 for the combination $\langle i,j \rangle$; probability $p_i$ for other combinations with $i$ in cell 1; probability $p_j$ for other combinations with $j$ in cell 2; and probability $p_0$ for all combinations involving neither $i$ nor $j$, where the values of $p_i, p_j, p_0$ are listed in (10) and their relative magnitudes are described in (11).\footnote{The expressions in (10) are combinatorial expressions which can be described as follows. Each expression tallies the possible alternatives for freely placing some number of `1's in a $m \times n$ grid, given that either one or two `1's have already been placed in the grid, such that the final arrangement contains a total of $k$ `1's in the $m \times n$ grid squares.

\hspace{0.5cm} Specifically, in (10): $N$ is the number of ways of placing $k-1$ instances of `1', given that one instance is already placed at position $\langle i,j \rangle$. $p_i$ is the number of ways of placing $k-2$ instances of `1', given that one instance is already placed at position $\langle i,j \rangle$ and one is placed elsewhere in row $i$. $p_j$ is the number of ways of placing $k-2$ instances of `1', given that one instance is already placed at position $\langle i,j \rangle$ and one is placed elsewhere in column $j$. $p_0$ is the number of ways of placing $k-2$ instances of `1', given that one instance is already placed at position $\langle i,j \rangle$ and one is placed elsewhere, away from row $i$ and column $j$. 

\hspace{0.5cm}  The calculation of each of these quantities is complicated by the fact that we only want to count arrangements of `1's that ensure that every row and every column contains at least one instance. To perform the calculation, we need to consider a more general problem: placing the `1's such that we fill all rows and columns except for $r$ not-to-be filled rows, and $s$ not-to-be-filled columns. We start with $r=0$ and $s=0$, which is the special case that we actually want to solve. The fact that we start from here explains why the sums in each of the expressions in (10) start from $r=0$ and $s=0$. The sub-expressions such as $\binom{m-1}{r}$ count the number of ways of selecting the $r$ exceptional, not-to-be-filled rows, and $\binom{n-1}{s}$ count the number of ways of selecting the $s$ exceptional, not-to-be-filled columns. (The number of options available for selecting the not-to-be-filled rows and columns will depend on how many rows and columns have already had a `1' assigned to them \textit{a priori} in the definitions of $N$, $p_i$, $p_j$ and $p_0$, and hence these sub-expressions differ slightly among the expression in (10)). The rightmost pieces of the expressions in (10) count all the number of ways to place the set of `1's in the available grid squares of the non-exceptional rows and columns. Now, consider that we start with $r=0$ and $s=0$; we have excluded no rows or columns and we simply distribute our `1's anywhere we can, and we count the number of possibilities. However, some of those distributions will, purely by accident, have left exactly one or more rows empty: this means we have over-counted, since we don't want any rows empty. To correct for the over-counting, we first calculate the number of ways of distributing the `1's with one row explicitly excluded (i.e., we set $r=1$), and we \textit{subtract} this from our first count. But in doing this, we have also subtracted away arrangements which by accident left two rows empty, and moreover we have subtracted them not once but twice, since each such pair of empty rows is counted once as (row$_1$ = deliberately empty, row$_2$ = accidentally empty) and once as (row$_1$ = accidentally empty, row$_2$ = deliberately empty). Thus, we need to compensate for the over-subtraction by \textit{adding} the number of ways of distributing the `1's with two rows explicitly excluded (i.e., $r=2$); and in a similar fashion this will give rise to an over-counting of arrangements that have three rows empty, and so on. The right answer, therefore, is obtained by alternatively adding and subtracting the ways of counting arrangements with $r$ excluded rows and/or $s$ excluded columns (i.e., following a widely-used method known as the inclusion-exclusion principle), and this is why each of the expressions in (10) begin with sums that start at $r=0$ and $s=0$ and count upwards (as the number of excluded rows and columns ticks upwards), while the expression $(-1)^{r+s}$ causes the flipping back and forth of addition and subtraction.}

\begin{table}[H]
\begin{tabular}{cc|c|c|c|c|c|c|c|c|c|}
\cline{3-11}
(9)                   &          & $1$      & $2$      & $\cdots$  & $(j-1)$  & $j$      & ($j+1)$  & $\cdots$  & $(n-1)$  & $n$      \\ \cline{2-11} 
\multicolumn{1}{c|}{} & $1$      & $p_0$    & $p_0$    & $\cdots$  & $p_0$    & $p_j$    & $p_0$    & $\cdots$  & $p_0$    & $p_0$    \\ \cline{2-11} 
\multicolumn{1}{c|}{} & $2$      & $p_0$    & $p_0$    & $\cdots$  & $p_0$    & $p_j$    & $p_0$    & $\cdots$  & $p_0$    & $p_0$    \\ \cline{2-11} 
\multicolumn{1}{c|}{} & $\vdots$ & $\vdots$ & $\vdots$ & $\ddots$  & $\vdots$ & $\vdots$ & $\vdots$ & $\iddots$ & $\vdots$ & $\vdots$ \\ \cline{2-11} 
\multicolumn{1}{c|}{} & $(i-1)$  & $p_0$    & $p_0$    & $\cdots$  & $p_0$    & $p_j$    & $p_0$    & $\cdots$  & $p_0$    & $p_0$    \\ \cline{2-11} 
\multicolumn{1}{c|}{} & $i$      & $p_i$    & $p_i$    & $\cdots$  & $p_i$    & 1        & $p_i$    & $\cdots$  & $p_i$    & $p_i$    \\ \cline{2-11} 
\multicolumn{1}{c|}{} & $(i+1)$  & $p_0$    & $p_0$    & $\cdots$  & $p_0$    & $p_j$    & $p_0$    & $\cdots$  & $p_0$    & $p_0$    \\ \cline{2-11} 
\multicolumn{1}{c|}{} & $\vdots$ & $\vdots$ & $\vdots$ & $\iddots$ & $\vdots$ & $\vdots$ & $\vdots$ & $\ddots$  & $\vdots$ & $\vdots$ \\ \cline{2-11} 
\multicolumn{1}{c|}{} & $(m-1)$  & $p_0$    & $p_0$    & $\cdots$  & $p_0$    & $p_j$    & $p_0$    & $\cdots$  & $p_0$    & $p_0$    \\ \cline{2-11} 
\multicolumn{1}{c|}{} & $m$      & $p_0$    & $p_0$    & $\cdots$  & $p_0$    & $p_j$    & $p_0$    & $\cdots$  & $p_0$    & $p_0$    \\ \cline{2-11} 
\end{tabular}
\end{table}

\begin{table}[H]
\begin{tabular}{lrll}
(10) & $p_i$     & = & $\frac{1}{N} \sum\limits_{r=0}^{m-1} \sum\limits_{s=0}^{n-2} (-1)^{r+s} {\binom{m-1}{r}} {\binom{n-2}{s}} {\binom{(m-r)(n-s)-2}{k-2}}$ \\
     & $p_j$     & = & $\frac{1}{N} \sum\limits_{r=0}^{m-2} \sum\limits_{s=0}^{n-1} (-1)^{r+s} {\binom{m-2}{r}} {\binom{n-1}{s}} {\binom{(m-r)(n-s)-2}{k-2}}$ \\
     & $p_0$     & = & $\frac{1}{N} \sum\limits_{r=0}^{m-2} \sum\limits_{s=0}^{n-2} (-1)^{r+s} {\binom{m-2}{r}} {\binom{n-2}{s}} {\binom{(m-r)(n-s)-2}{k-2}}$ \\
     & where $N$ & = & $\sum\limits_{r=0}^{m-1} \sum\limits_{s=0}^{n-1} (-1)^{r+s} {\binom{m-1}{r}} {\binom{n-1}{s}} {\binom{(m-r)(n-s)-1}{k-1}}$            
\end{tabular}
\end{table}

\begin{table}[H]
\begin{tabular}{llllll}
(11) & $1$ & $>$ & $p_0$ & $>$ & $p_i, p_j$ 
\end{tabular}
\end{table}

Contingency table (9), taken together with the inequalities in (11), tells us that under \textit{any} conditions, if the evidence lexeme and the focal lexeme are the same in cell 1 (corresponding to row $i$ in table (9)), then the possibility with the highest probability is that they are the same in cell 2 (column $j$ in table (9)); and if the evidence lexeme and the focal lexeme are different in cell 1 (corresponding to all other rows in table (9)), then the possibilities with the equal-highest probabilities are that they are different in cell 2 (all columns other than $j$). For a reasoner attempting to maximise its chances of guessing correctly, the heuristic under all conditions is that same implies same; different implies different. 

We would emphasise that the result we have arrived at is not caused by learning biases\footnote{We thank an anonymous reviewer for asking us about this.}---since the reasoning process here involves no learning---but is due solely to the application of rational, probabilistic inference within the confines of a four-part analogical reasoning task. Also, we do not presuppose that speakers actually calculate these probabilities. Rather, what our demonstration shows is that in order to act rationally, all a speaker needs to do is to follow the dictum that same implies same and different implies different; calculating actual probabilities would not deliver any further benefit.\footnote{We leave for future studies the question of what would happen if the reasoner, instead of merely maximising their chance of guessing right, attempted to match the relative probabilities in table (9). We note that in order to do so, the reasoner would need access to probabilities $p_i$, $p_j$ and $p_0$, which are sensitive to the cell-specific values of $m$, $n$ and $k$, rather than just relying on an invariant, simple heuristic. Consequently, the probability-matching strategy would presumably be significantly more cognitively demanding than the probability-maximising one which we consider here.} Lastly, we note that `same--same; different--different' heuristics also arise elsewhere in human cognitive behaviour, such as in backward blocking \citep{shanks1985forward,sobel2004children} and in referent selection during fast mapping \citep{carey1978acquiring, horst2008fast}.

\subsection{A revised paradigm cell filling process}\label{PCF-revised}

\begin{figure}[ht]
\caption[Paradigm cell filling mechanism]{Paradigm cell filling mechanism with associative evidence and dissociative evidence. (a) Select a pivot cell and examine its contents in all lexemes. (b) `Matching' lexemes are those whose pivot cell matches that of the focal lexeme; others are `contrasting'. (c) Examine the focal cell contents. (d) Exponents score +1 for each matching lexeme, $-\alpha$ for each contrasting lexeme; select the highest-scoring exponent.}
\centering
\includegraphics[width=9cm]{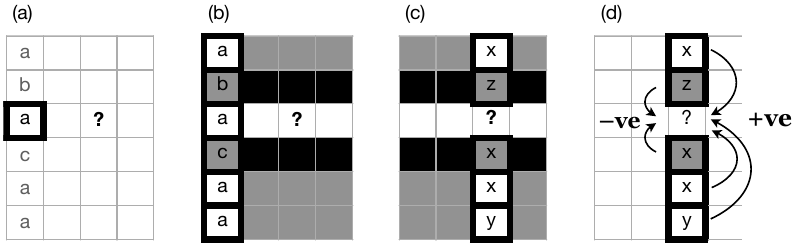}
\label{pcfp-positive-negative}
\end{figure}

The heuristic `same implies same; different implies different' can be integrated into our inflectional model by revising the paradigm cell filling process. Accordingly, Figure~\ref{pcfp-positive-negative} schematises a paradigm cell filling process which attends not only to associative evidence (`same implies same') but also to dissociative evidence (`different implies different'). As before (cf Figure~\ref{pcfp-positive-only}), the model randomly selects a pivot cell as a basis for inference, and notes the exponent of the pivot cell in the focal lexeme (stage (a) of Figure~\ref{pcfp-positive-negative}). The model then scans the pivot cells of all other lexemes to gather evidence, but instead of selecting only those lexemes whose pivot exponent matches the pivot exponent of the focal lexeme, it selects all lexemes, classifying them as either `matching' lexemes (shaded grey in Figure~\ref{pcfp-positive-negative}(b)) or ‘contrasting' lexemes (shaded black). Next, the model notes the exponents of the focal cells in both kinds of evidence lexeme. Exponents of matching evidence lexemes are counted positively; exponents of contrasting evidence lexemes are counted negatively; and the exponent with the highest score is selected for the focal exponent. In our model implementation, we score a token of an exponent in a matching evidence lexeme as +1 and a token of an exponent in a contrastive evidence lexeme as $-\alpha$, where $\alpha$ is a non-negative parameter whose value is set by the experimenter. Setting $\alpha = 0$ yields a model like those in Sections \ref{existing} and \ref{family}, which solely attends to associative evidence; setting it greater than zero brings both associative evidence and dissociative evidence into play.

\begin{figure}[h!t]
\caption{Sampling 20\% of evidence lexemes, using 4 pivots, incorporating dissociative evidence ($\alpha=0.1$). 20 simulations of 20,000 cycles.}
\centering
\includegraphics[width=10.5cm]{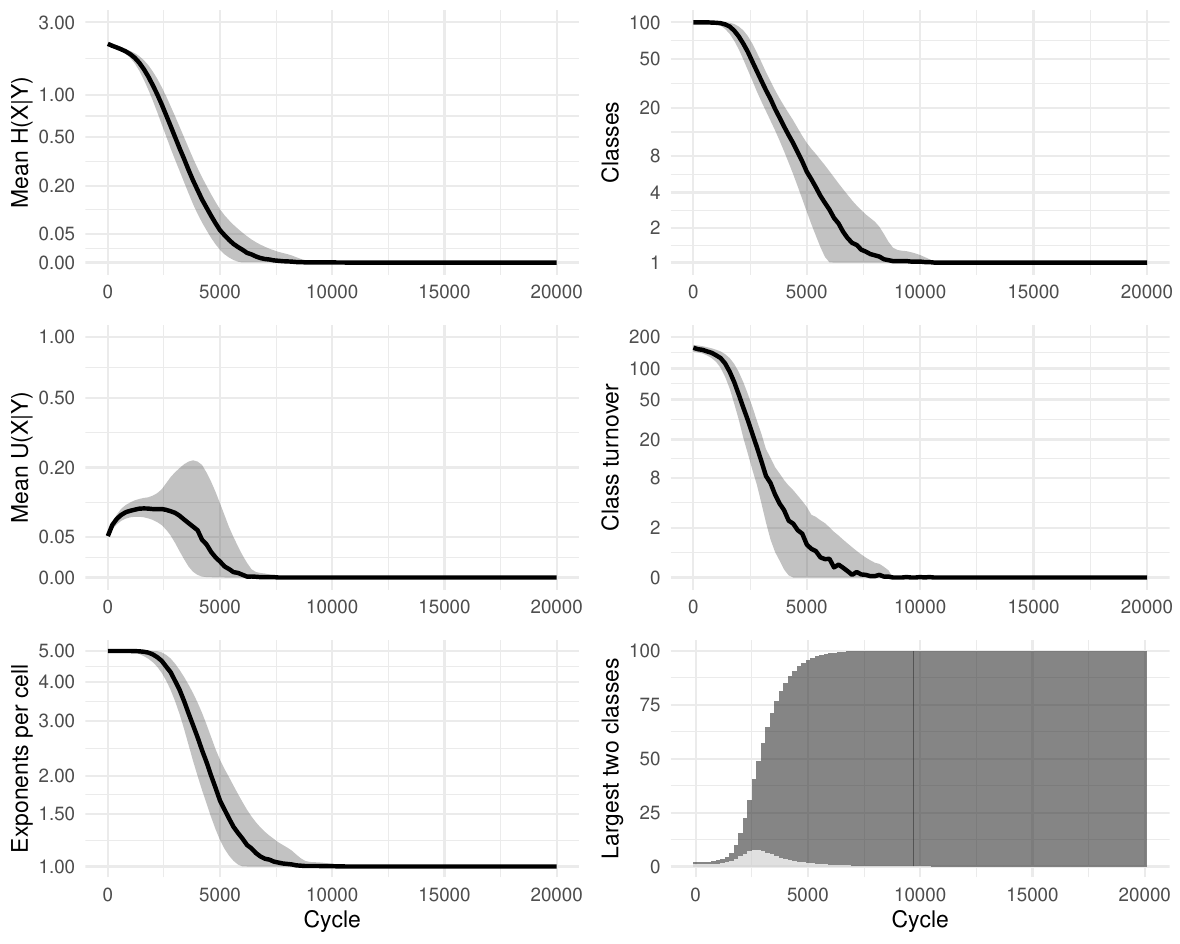}
\label{plot-NE-alpha-01}
\end{figure}

\begin{figure}[ht!]
\caption{Sampling 20\% of evidence lexemes, using 4 pivots, incorporating dissociative evidence ($\alpha=0.2$). 20 simulations of 20,000 cycles.}
\centering
\includegraphics[width=10.5cm]{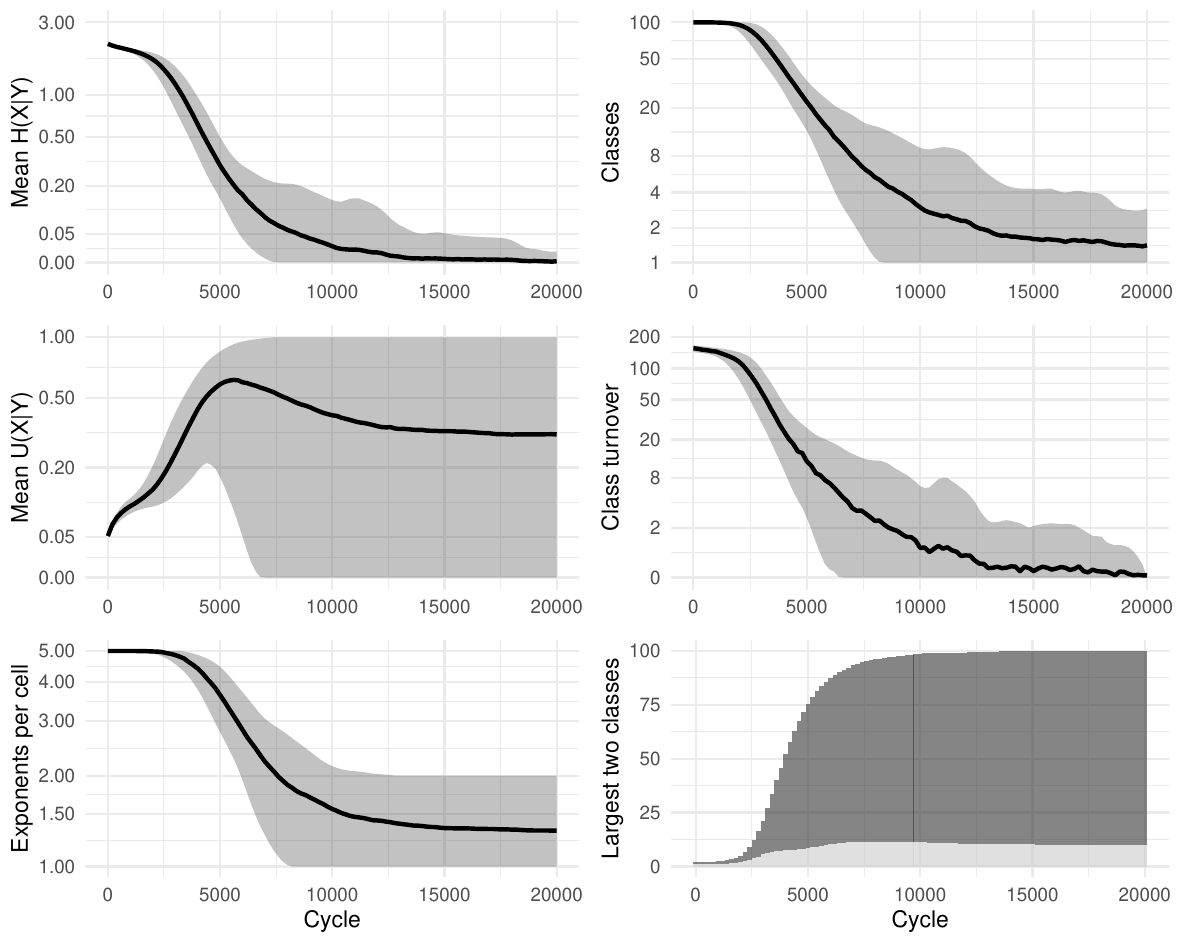}
\label{plot-NE-alpha-02}
\end{figure}

\begin{figure}[ht!]
\caption{Sampling 20\% of evidence lexemes, using 4 pivots, incorporating dissociative evidence ($\alpha=0.5$). 20 simulations of 50,000 cycles.}
\centering
\includegraphics[width=10.5cm]{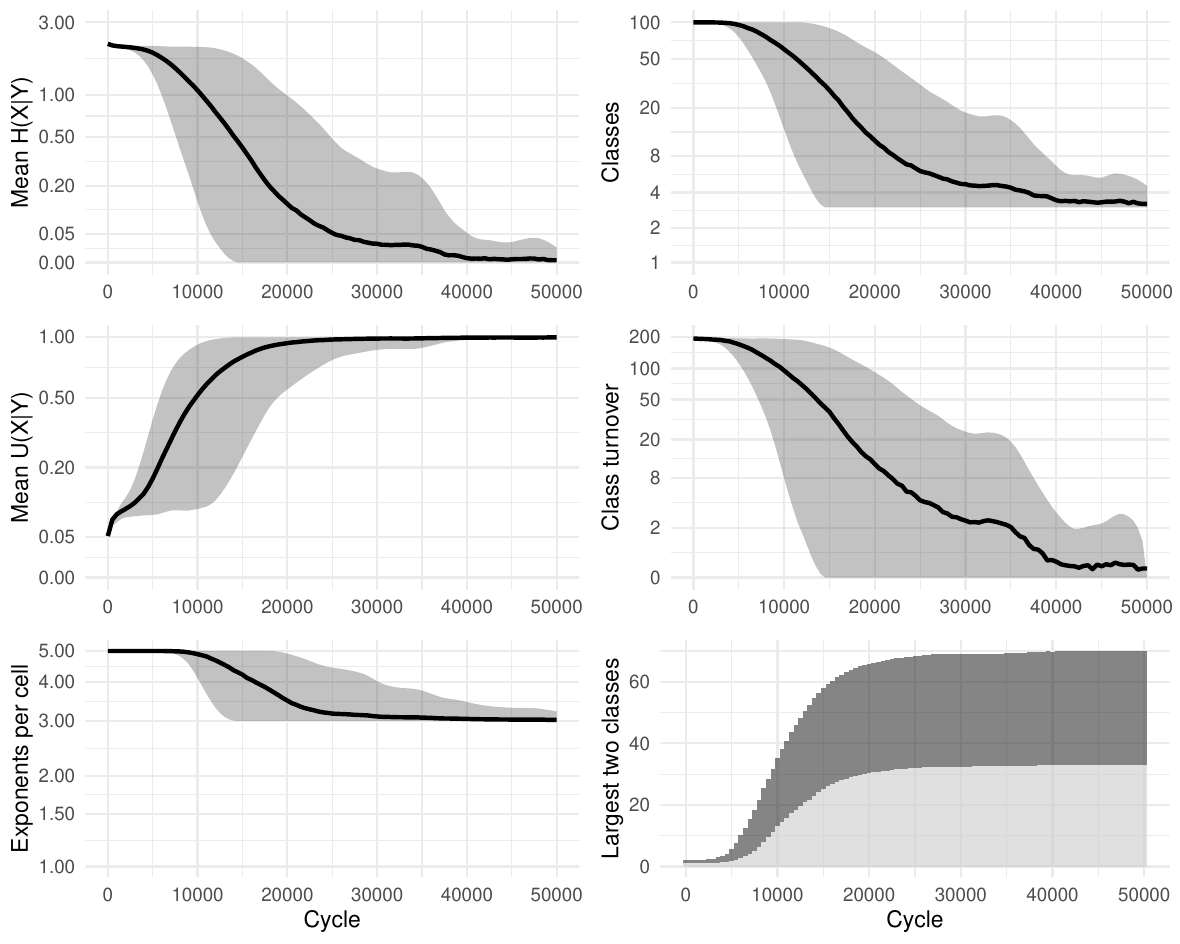}
\label{plot-NE-alpha-05}
\end{figure}

\begin{figure}[ht]
\caption{Eight snapshots evenly spaced between cycle 0 (leftmost) and cycle 50,000 (rightmost) from one simulation incorporating dissociative evidence ($\alpha=0.5$). Each snapshot shows 100 lexemes in rows, 8 cells in columns. Distinct exponents in each cell are indicated by shading. Lexemes have been ordered vertically so as to accentuate the three final inflection classes most clearly. A fourth class can be seen collapsing between snapshots 5 and 6.}
\centering
\includegraphics[width=12cm]{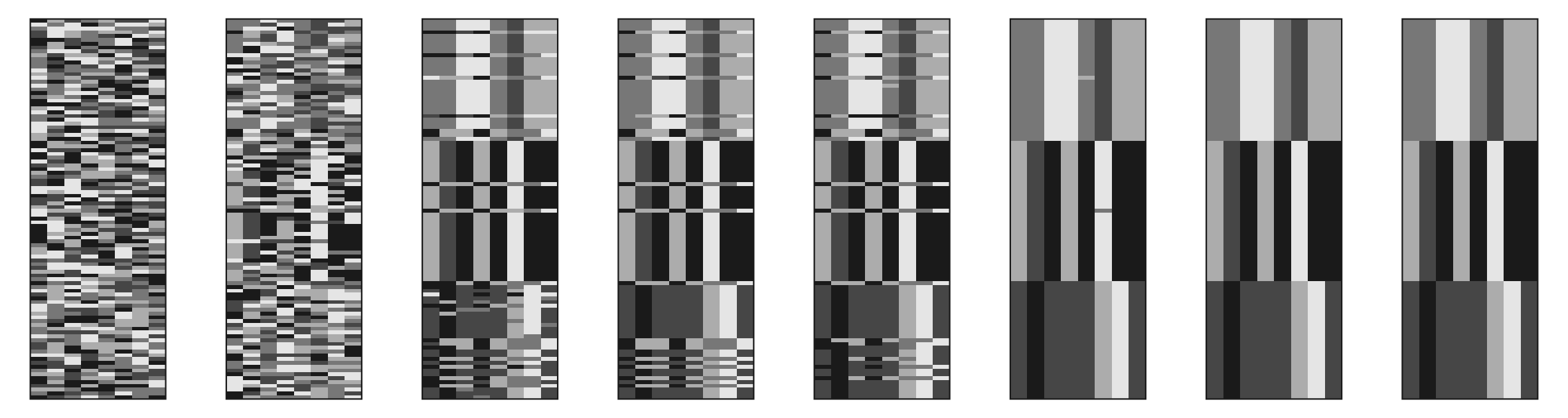}
\label{snapshots-alpha05}
\end{figure}

\begin{figure}[ht!]
\caption{Sampling 20\% of evidence lexemes, using 4 pivots, incorporating dissociative evidence ($\alpha=0.67$). 20 simulations of 100,000 cycles.}
\centering
\includegraphics[width=10.5cm]{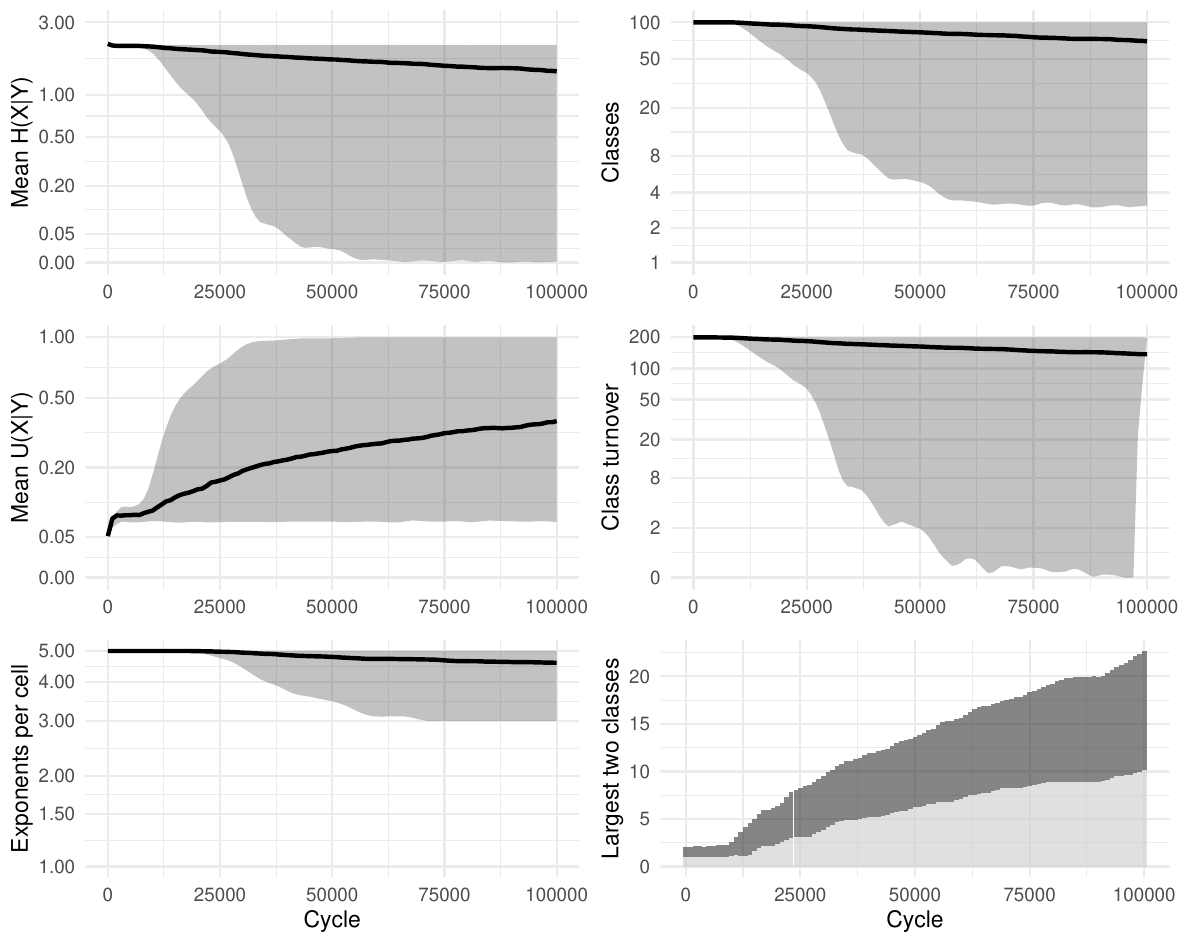}
\label{plot-NE-alpha-067}
\end{figure}

In a model as abstract as ours, we caution against attributing any overly concrete interpretation to $\alpha$; it is simply a device by which we can raise or lower the relative contribution of dissociative evidence and observe the consequences. In Figures \ref{plot-NE-alpha-01}--\ref{plot-NE-alpha-067}, we show the consequences of incorporating dissociative evidence. For brevity, we focus here on the emergence of inflection classes (rhizomorphomes), however see Appendix A for corresponding models and results for the emergence of metamorphomes. In all cases, we use 4 pivots and 20\% of the evidence lexemes, sampled with a uniform (non-Zipfian) distribution. Having examined other settings \citep[see][]{round2022cognition}, we find that they interact with $\alpha$ in non-linear ways that merit examination in future studies. Here we focus on highlighting the various outcomes that can result once dissociative evidence is attended to, and in particular the consequences for stable autonomous morphology.

For values of $\alpha$ close to zero (Figure~\ref{plot-NE-alpha-01}), where associative evidence greatly outweighs dissociative evidence, the system is quickly dominated by a single inflection class, echoing the behaviour we saw in associative-only models in Sections \ref{existing} and \ref{family}.

For the sampling settings we are using, the first qualitatively different outcome occurs at around $\alpha=0.2$, i.e., one-fifth the strength of associative evidence (Figure~\ref{plot-NE-alpha-02}). Here, some runs of the simulation still collapse to a single class, but most preserve two distinct classes, and most cells preserve a two-way contrast in their exponents. In striking difference to the associative-only models, Theil's U climbs to a value close to $1$, indicating that cells have become strongly predictive of one another. Mean conditional entropy sinks dramatically but without reaching zero, instead settling at level which is low, yet positive. The turnover metric shows that until 20,000 cycles, some amount of change continues to occur in the precise exponents that comprise the inflection classes, with new classes occasionally innovated, but thereafter the system settles into complete stability. At $\alpha=0.5$, i.e., half the strength of associative evidence (Figure~\ref{plot-NE-alpha-05}), the system evolves similarly, this time settling into a state with three, stable classes after around 40,000 cycles. Figure~\ref{snapshots-alpha05} shows evenly-spaced snapshots illustrating one of the simulations in which four distinct classes emerge.

A second qualitative change in behavior occurs not far above $\alpha=0.5$. At $\alpha=0.67$, i.e., two-thirds the strength of associative evidence, the system remains highly agitated and retains its initial 100 classes for 25,000 cycles. By 100,000 cycles, coherence is still only just beginning to form: the two largest classes now cover on average a little less than half the lexicon, while the remainder comprises a large number of tiny classes. Nevertheless, as we saw at lower $\alpha$ values, Theil's U is climbing, revealing an emergent self-organisation that is built on mutual predictiveness between cells, not the elimination of diversity. At $\alpha=0.75$, i.e., three-quarters the strength of associative evidence, the system simply remains chaotic, with not even a single class able to secure more than one member lexeme after 100,000 cycles.

The incorporation of dissociative evidence introduces a fundamental change to the evolutionary dynamics of the models. In addition to the attraction dynamic of the models discussed in Sections \ref{existing} and \ref{family}, which promotes pure convergence of rhizomorphomes and metamorphomes, models that incorporate dissociative evidence also introduce a repulsion dynamic which promotes divergence. While the attraction dynamic still causes local, similar clusters of lexemes (or cells) to coalesce into single classes, the repulsion dynamic pushes those local clusters further away from one another. If the repulsive force is overly strong, it prevents even local clusters from coalescing. If it is too weak, it is overwhelmed by the attraction force and the various local clusters merge and collapse into a single class. At an intermediate strength, it is weak enough to enable local clusters to coalesce, but also strong enough to prevent the system from collapsing entirely into one class; the result is a stable dynamic regime \citep{lorenz1963deterministic, holling1973resilience, Mayer2004regime} with multiple resilient classes, reminiscent of what is observed in natural language.

Our opening  question was, how can autonomous morphology persist over time? The `how possibly' answer that we offer here is that dissociative evidence is available to the rational reasoner. As our models demonstrate, when analogical inference makes use of dissociative evidence, then under the right conditions, the outcome is an attraction-repulsion dynamic in which multiple classes can arise and, more importantly, persist. The parallels between this outcome and the properties of naturally occurring inflectional systems furnish compelling reasons to entertain the following hypothesis: morphomic structure is a form of inflectional self-organisation, liable to emerge spontaneously via a simple inferential process that attends to both associative evidence and dissociative evidence.

\FloatBarrier
\section{Discussion}\label{discussion}

Our simulations demonstrate that the introduction of a moderately strong repulsion dynamic leads to models developing and maintaining stable paradigmatic structure corresponding to the morphomic structures observed in natural language: inflection classes (rhizomorphomes), and recurrent groupings of paradigm cells (metamorphomes). Indeed, it is only the introduction of such a dynamic that consistently promotes the emergence of structured variation. Since attraction-only models (in which lexemes only grow more similar to each other) inevitably remove all variation, they cannot evolve the stable, structured diversity characteristic of inflectional systems. By contrast, models with an attraction-repulsion dynamic create conditions in which morphome-like structure emerges spontaneously and endures over long periods. Thus, just one small ingredient---rationally-motivated change based on dissimilarity---separates models that tend inexorably to uniformity from models which evolve resilient, morphome-like structure.

Working backwards, by comparing the results of our simulations with the scenarios observed in natural language, we can first discriminate between plausible dynamics (those which produce empirically recognisable characteristics) and implausible dynamics (those which produce empirically unattested characteristics). On this basis, we can distinguish between model conditions which promote plausible dynamics and model conditions which promote implausible dynamics, and ultimately between the hypotheses about inflectional systems which underlie each set of conditions. Working forwards, we observe how apparently simple processes can lead to the emergence of diverse, complex systems, and, moreover, to considerable longevity of those systems once established. Via this process of exploratory testing and comparison, models such as ours have the potential to alter how morphologists and historical linguists approach the task of accounting for inflectional complexity.

Our results mount a substantive empirical challenge to the view that autonomous morphology is `unnatural' \citep{Aronoff1994,KoontzGarboden2016}. In Section \ref{autonomous} we observed that the central explanatory challenge of autonomous morphology is to account for why the mean conditional entropy of inflectional systems typically remains stable at a low but non-zero level. The attraction-repulsion dynamic and the inferential mechanism underlying it which we have identified here provide such an account. In Section \ref{autonomous} we also questioned the assumption that autonomous morphology is inherently `unnatural'. Initial reasons to doubt this assumption include the facts that autonomous morphology is cross-linguistically common \citep{Bach2020,Herce2020}, productively replicated by speakers, and resilient over long time periods \citep{Maiden2001,Maiden2003,Maiden2005,Maiden2013,Maiden2016,Maiden2018,Enger2014,Esher2017,Esher2023,FeistPalancar2021,Esher2022b,cappellaro2023cognitive}. Our study affords a deeper insight into the issue via a principled exploration of the systemic properties which give rise to inflectional organisation (or lack thereof). Where descriptive linguistics shows that autonomous morphology is typologically natural (i.e., commonly observed), our model reveals that, at a causal level, autonomous morphology is the natural (i.e., automatic, emergent, even unavoidable) outcome of a dynamic system which changes according to a process of natural (i.e., rational) inference. In these multiple respects, morphomic structure is a fundamentally natural feature of human language; attempted characterisations of autonomous morphology as unnatural prove ill-founded and misleading. 

Having established that autonomous morphology is natural, a separate question is whether it is mentally represented \citep[see][on the psychological reality of morphomes]{Maiden2018,FeistPalancar2021}. Through computational modelling, we have demonstrated that morphome-like structure can emerge via a dynamic process consisting merely of piecemeal individual changes, and moreover, within a system which lacks any explicit representation of morphomic structure (e.g. by means of morphomic indices, such as inflection class indices or morphomic stem indices). Just as the architecture of \citeauthor{AckermanMalouf2015}’s model at no point made explicit reference to mean conditional entropy or to a goal of reducing mean conditional entropy, the input and internal mechanisms of our models make no explicit reference to morphomic structure: yet morphomic structure, like low mean conditional entropy, arises as an emergent property of the system and robustly persists thereafter. Given this result, it may be tempting to conclude that morphomes need not be mentally represented. However, a number of logical steps would need filling in first. Our model is deliberately idealised, and excludes such representations by design. In order to be convincing, a claim that the mind also lacks them would require at minimum an argument, or model comparisons, showing that there is no cognitive advantage to such representations for the speaker. Such demonstrations will almost surely be conditional on prior assumptions about language in the mind and therefore be substantially theory-dependent.\footnote{See \citet{ONeill2014} for some indications that morphomic structures may be represented without explicit indexing in a paradigmatic, abstractive approach. Constructive approaches, such as those of \citet{Aronoff1994} and to some extent \citet{Stump2015}, require explicit stipulation of morphomic distribution patterns.} Thus, we do not see our results as immediately resolving debates over representation in the same way that they bring resolution to the question of naturalness.

\section{Conclusions}\label{conclusions}

Autonomous morphological structure is robustly attested in natural language, yet explaining its diachronic persistence has remained a significant challenge. In this paper, we develop a family of computational iterative models where inflectional change is driven by a paradigm cell filling task, to investigate how a succession of simple analogical changes can lead to morphological self-organisation. By manipulating the constituent parameters of the models, we are able to probe the conditions under which stable morphomic structure arises or does not arise. Our study reveals for the first time some of the simplest conditions under which resilient inflection class systems and paradigmatic distribution patterns can potentially emerge.

Our study develops a range of simulations based on the seminal model of \citet{AckermanMalouf2015}. All the models iterate a paradigm cell filling task, though implementing different parameters and strategies. In Ackerman and Malouf’s original model, lexemes only ever changed by increasing their resemblance to other lexemes: the resulting evolutionary dynamic is one of pure attraction, with the inescapable outcome that all inflection classes eventually merge into a single class. An equivalent inbuilt structural problem is observed for the related model by \citet{Esher2015a,Esher2015b,Esher2017} for metamorphomes. Although both these models succeed in reducing the mean conditional entropy of the system, closer inspection showed that this results not from increasing mutual predictability between cells, but from reducing the variation within them, as our introduction of Theil’s U as a metric reveals. To the extent that these existing models exhibit self-organisation, it is only of a radically homogenising kind, which is not probative for the emergence of diachronically robust morphomic structure. Even when considerably enriched with more realistic frequency distributions and sampling methods, attraction-only models never evolve the persistent, structured diversity which characterises inflectional systems in natural language: indeed, they cannot do so, as the evolutionary dynamic inherent to these models inevitably removes all variation in favour of uniformity and zero entropy.

By contrast, attraction-repulsion models enable the emergence of stable variation, via the combination of pressure to coalesce, which pushes the system towards a lower number of variants, and pressure to disperse, which pushes the system to keep the remaining variants distinct. As the models run, the initially random lexicons progressively self-organise into inflection classes via the occurrence of individual analogical changes affecting single exponents, illustrating the systematicity and ease with which abstract morphological structure can arise spontaneously within an inflectional system. The key distinction drawn here between attraction-only and attraction-repulsion models resides in the nature of analogical reasoning during the paradigm cell filling task. In attraction-only models, analogy is based purely on `associative evidence’, whereas in our attraction-repulsion model, a moderate weight is also given to dissociative evidence, which we have argued is also available to a rational reasoner. Only attraction-repulsion models result in the emergence of stable inflection class systems and paradigmatic distribution patterns. Thus, our investigation identifies a plausible route by which stable morphomic structure may emerge: the crucial mechanism is a paradigm cell filling process in which inference is sensitive to dissociative evidence, a possibility which has largely been overlooked in prior research on inflectional analogy.

Secondly, our plausible route to emergent autonomous morphology is one which is intrinsically natural. It is based on a natural (i.e., rational) reasoning process which gives rise to a dynamical system whose natural (i.e., emergent) outcome is stable morphomic structure. Thus we argue that autonomous morphology is far from the `unnatural' phenomenon it has often been cast as, by its proponents and detractors alike. Autonomous morphology should instead be understood as a newly-identified kind of natural feature of human language.

Our simulations demonstrate that morphome-like structure can emerge via a dynamic process consisting merely of piecemeal individual changes. While the input and internal mechanisms of our models make no explicit reference to morphomic structure, such structure nevertheless arises as an emergent property of the system, and robustly persists once established. Given this finding, and the fact that autonomous morphology is widespread in human language, it is natural to ask next what form the cognitive representation of autonomous morphology might take. We caution that the deliberately idealised implementation of our current model admits multiple, often theory-dependent, interpretations regarding mental representations of inflection (see e.g. \citeauthor{Roundforthcoming}, forthcoming, for discussion and illustration); thus, at present, our results should not be taken to embody any particular claim about how inflectional structure, including morphomic structure, is represented in the mind. These remain intriguing topics of investigation for future research.

Finally, our study illustrates the promise and potential of computational evolutionary modelling to shed light on fundamental properties of human language. Computational models are particularly well suited for studying the outcomes of complex dynamical systems, where informal and intuitive arguments are unlikely to be reliable. Simple models are advantageous for their transparency: a small number of parameters can be varied independently and their effects studied, providing a clear interpretation and attribution of results. Conversely though, simple models may lack components (such as dissociative evidence) which are significant in real language, and so there is much to be gained from a careful process of `de-idealisation' \citep{mcmullin1985galilean}, adding parameters gradually and systematically. We expect that further de-idealisation of our own model will lead to additional insights. Promising directions include alternative implementations of dissociative evidence; more sophisticated internal representations; and multiple agents, equipped to reason using associative evidence and dissociative evidence, employing and transmitting morphological systems interactively.\footnote{For initial steps in this direction, see \citet{round2022cognition}.}

\FloatBarrier
\section*{Acknowledgements}

We would like to express thanks to colleagues who over several years have engaged us in helpful and insightful discussions at conferences and departmental seminars, at the 5th American International Morphology Meeting, online, 27–29 August 2021; 3rd International Symposium on Morphology, Toulouse, 22–24 September 2021; ‘Analogical Patterns in Inflectional Morphology’ conference, Berlin/online, 14 April 2022; Joint Conference on Language Evolution, Kanazawa/online, 5–8 September 2022; 26th International Conference on Historical Linguistics, Heidelberg, 4–8 September 2023; 4th International Symposium on Morphology, Nancy, 13–15 September 2023; Annual Conference of the Australian Linguistics Society, 30 November 2023; the Department of Human Behavior, Ecology and Culture, MPI-EVA, Leipzig, 21 January 2020; Laboratoire de Linguistique Formelle and Labex Empirical Foundations of Linguistics, Paris, 14 April 2023; Department of Linguistics, Macquarie University, 23 June 2023; Department of Comparative Language Science, Zurich, 27 February 2024; the Surrey Morphology Group, 22 November, 2023. We thank the editor and two anonymous reviewers for helpful comments and suggestions on the submitted manuscript. All shortcomings remain ours. We gratefully acknowledge funding that has supported this research, from: the Max Planck Institute for the Science of Human History for a research visit to Jena by Esher, 1–4  December 2019, during which we launched this collaborative project; Leverhulme Early Career Fellowship (ECF-2022-286) to Beniamine; British Academy Newton International Fellowship (NIF23 \textbackslash 100218) to Beniamine; Labex EFL International Chair 2023 to Round; British Academy Global Professorship (GP300169) to Round; and Horizon Europe Guarantee Consolidator Grant (EP/Y02429X/1) selected by the ERC and funded by the UKRI, to Round.

\FloatBarrier
\begin{appendices}

\section{Results of metamorphome simulations}\label{appendix-meta}

In the main paper we primarily reported simulations of evolving rhizomorphomes, i.e., inflection classes. Here we report corresponding results for metamorphomes. We use the same initial lexicon comprising 100 lexemes and 8 cells, with 5 allomorph indices distributed at random. In the metamorphomic model (Figure~\ref{pcfp-Esher}), the roles of lexemes and cells are reversed, which entails that we shift from a system with 8 possible pivot cells to one with 100 possible pivot lexemes, and from a system with 100 possible evidence lexemes to one with just 8 possible evidence cells.

\begin{figure}[ht!]
\caption{Sampling 1 pivot lexeme from a uniform distribution and 4 evidence cells. 100 simulations of 10,000 cycles. `Morphomic zones' are sets of cells which, in every lexeme, share their index.}
\centering
\includegraphics[width=10.5cm]{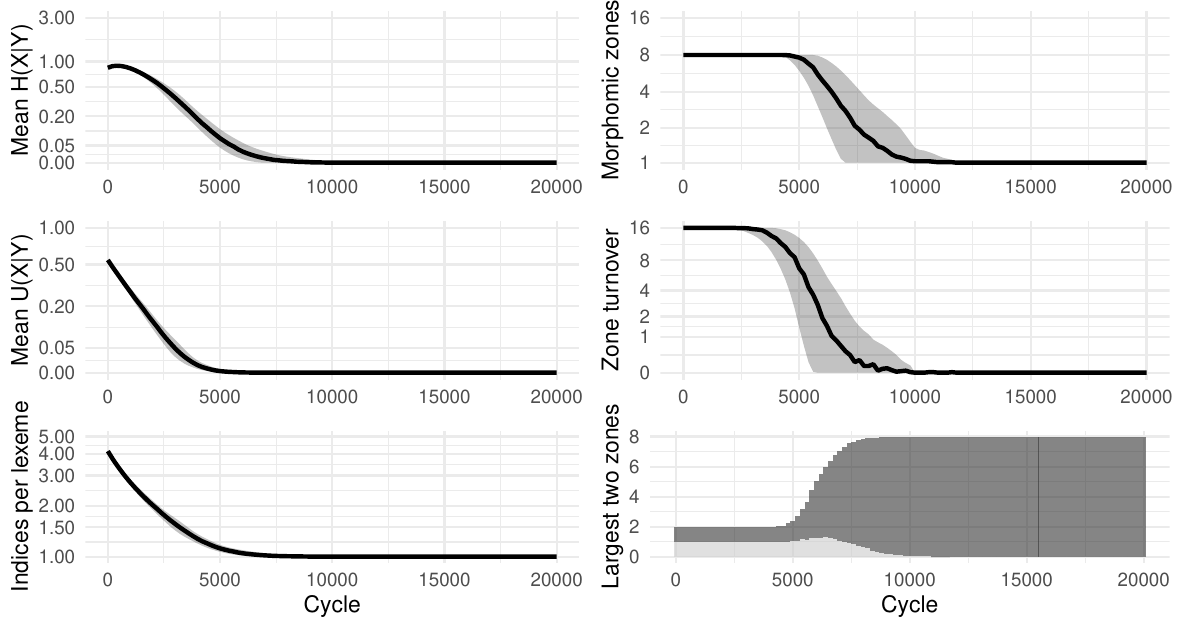}
\label{plot-meta-more-evidence}
\end{figure}

\begin{figure}[ht!]
\caption{Sampling 20\% of pivot lexemes and 1 evidence cell. 100 simulations of 30,000 cycles.}
\centering
\includegraphics[width=10.5cm]{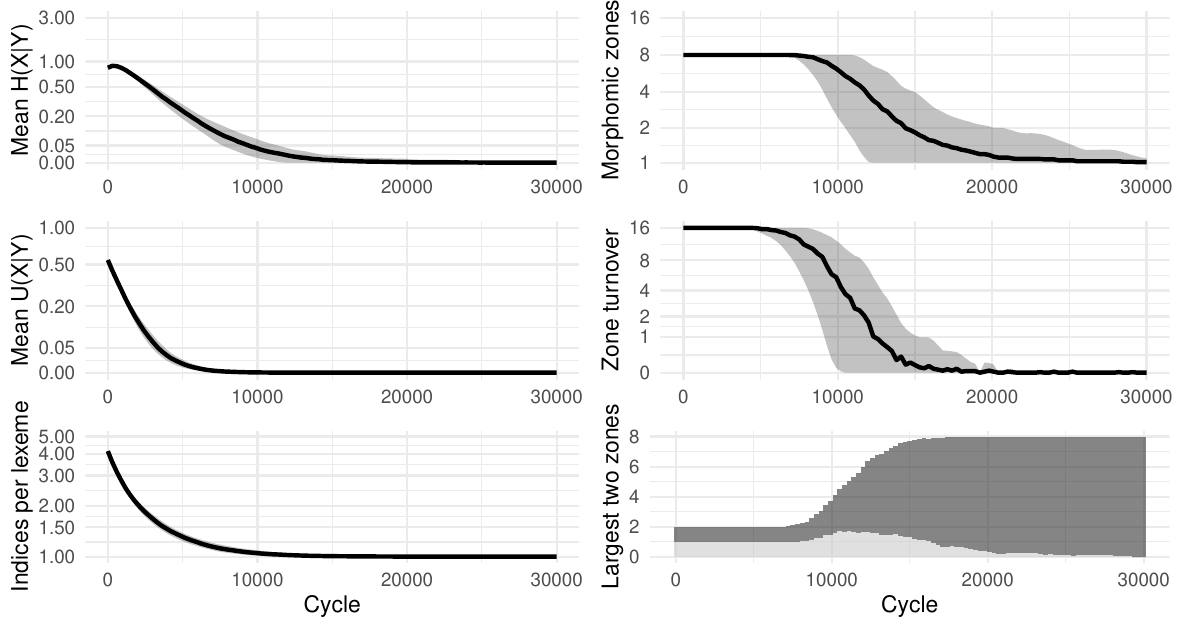}
\label{plot-meta-more-pivots}
\end{figure}

\begin{figure}[ht]
\caption{Eight snapshots evenly spaced between cycle 0 (leftmost) and cycle 30,000 (rightmost) from one simulation of \citeauthor{Esher2015a}'s (\citeyear{Esher2015a,Esher2015b}) model. Each snapshot shows 100 lexemes in rows, 8 cells in columns. Distinct allomorph indices in each lexeme are indicated by shading.}
\centering
\includegraphics[width=12cm]{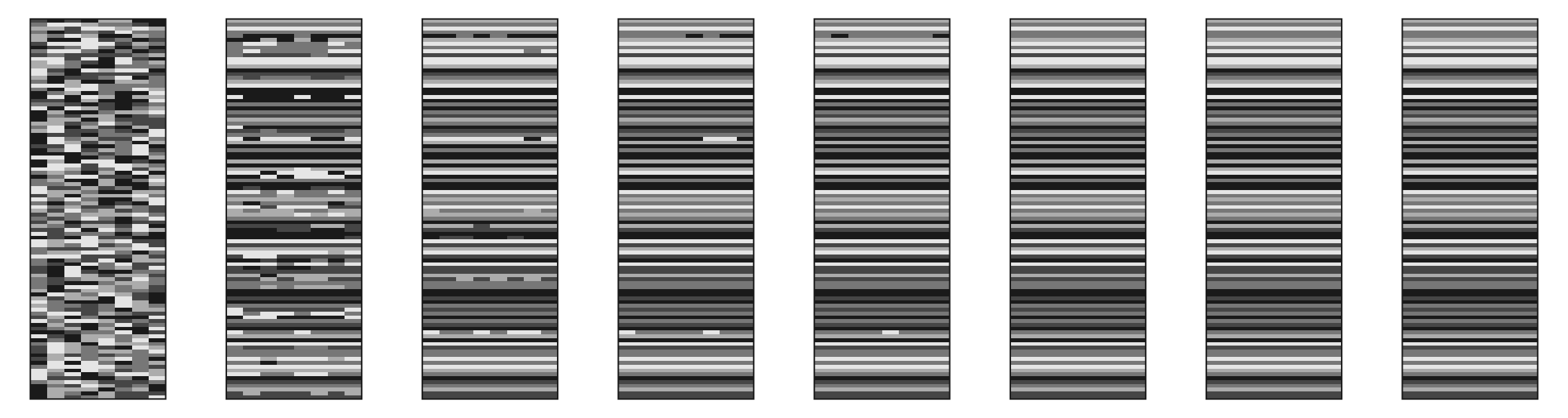}
\label{snapshots-appendix1}
\end{figure}

Esher's model sampled 1 pivot lexeme and 1 evidence cell using a uniform distribution. Figures~\ref{plot-meta-more-evidence} and \ref{plot-meta-more-pivots} increase this to 4 evidence cells and 20 pivot lexemes respectively. As in Section~\ref{family} for rhizomorphomes, adding extra evidence here does not result in the emergence of stable metamorphomic structure; rather, the metamorphomes soon collapse and $U(X\vert Y)$ falls to zero. Figure~\ref{snapshots-appendix1} shows snapshots from one of the simulations using 20 pivot lexemes, illustrating the shift of the lexicon from randomness to uniformity.

\begin{figure}[ht!]
\caption{Sampling 1 pivot lexeme from a uniform distribution and 1 evidence cell from a Zipfian distribution. 100 simulations of 50,000 cycles.}
\centering
\includegraphics[width=10.5cm]{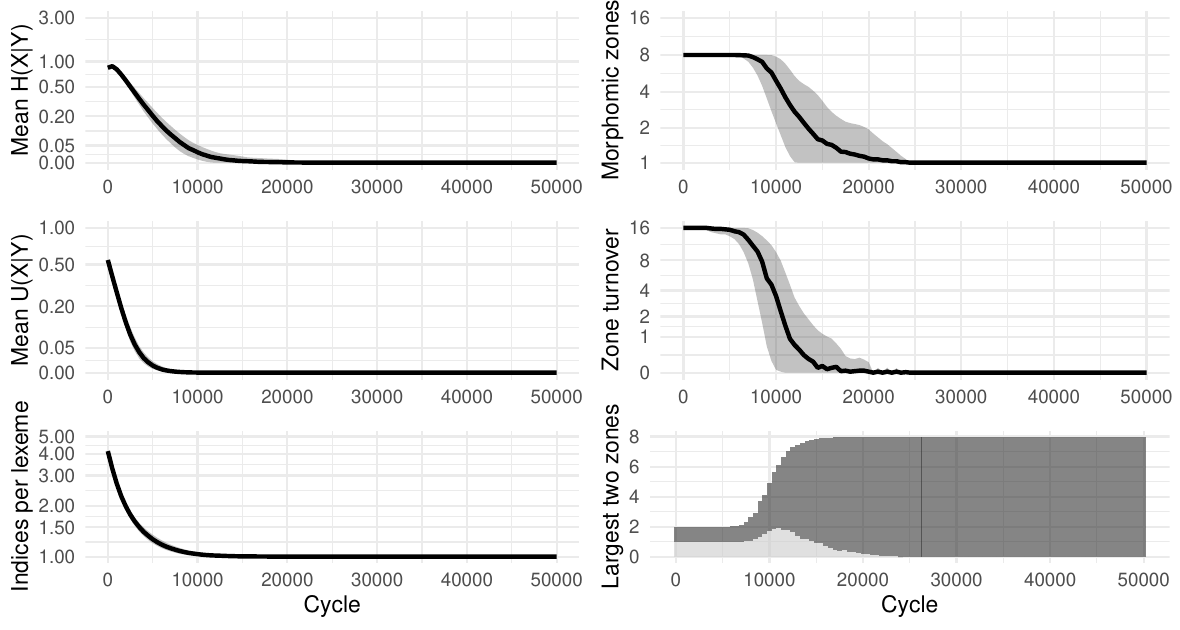}
\label{plot-meta-zipf-cells}
\end{figure}

\begin{figure}[ht!]
\caption{Sampling 10\% of pivot lexemes from a Zipfian distribution and 6 evidence cells from a uniform distribution. 100 simulations of 100,000 cycles.}
\centering
\includegraphics[width=10.5cm]{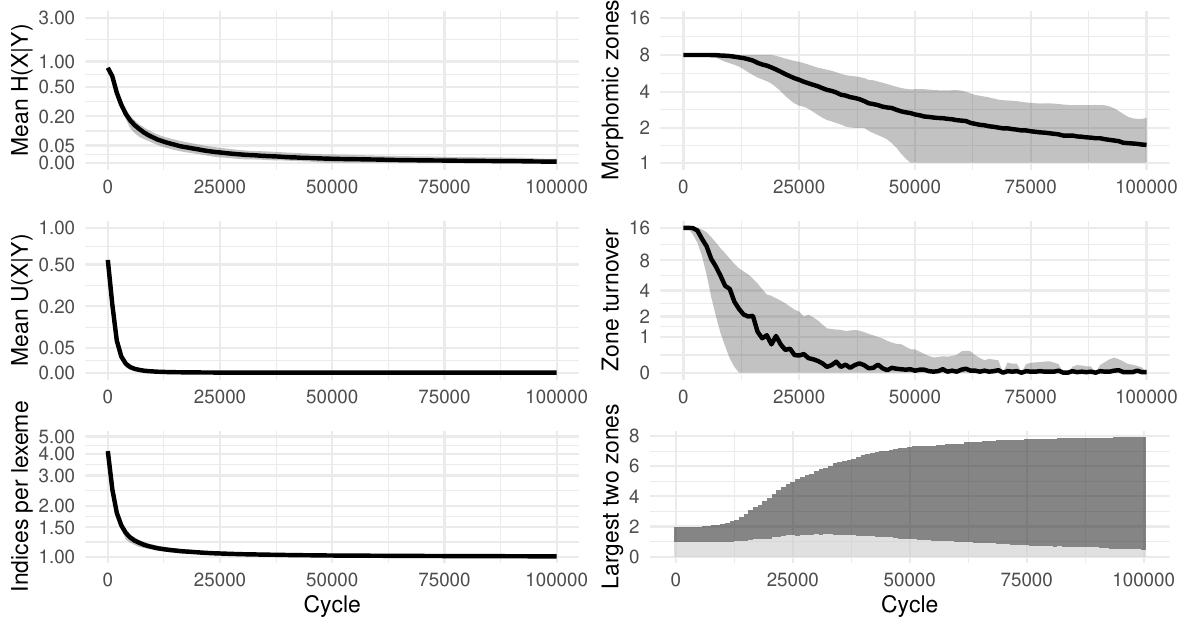}
\label{plot-meta-zipf-lexemes}
\end{figure}

Figures \ref{plot-meta-zipf-cells} and \ref{plot-meta-zipf-lexemes} respectively sample cells and lexemes using a Zipfian rather than uniform distribution. As we saw for rhizomorphomes in Section \ref{family}, adding Zipfian sampling does not result in the emergence of stable morphomic categories. Rather, since it causes certain cells and lexemes to change only very rarely, the effect of Zipfian sampling is to prolong the last stages of the system's trajectory, in which the final cells and lexemes succumb to a pattern which has attained dominance over the rest of the system.

\begin{figure}[ht!]
\caption{Sampling 10\% of pivot lexemes and 6 evidence cells from a uniform distribution, incorporating dissociative evidence ($\alpha = 0.25$). 100 simulations of 20,000 cycles.}
\centering
\includegraphics[width=10.5cm]{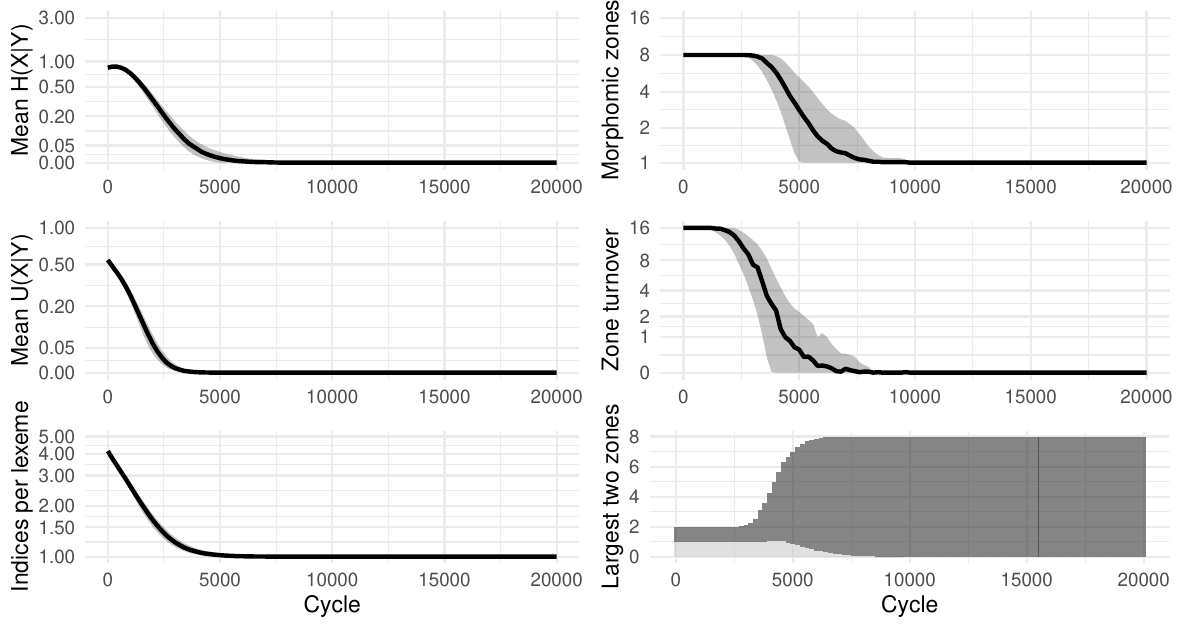}
\label{plot-meta-alpha025}
\end{figure}

\begin{figure}[ht!]
\caption{Sampling 10\% of pivot lexemes and 6 evidence cells from a uniform distribution, incorporating dissociative evidence ($\alpha = 0.5$). 100 simulations of 20,000 cycles.}
\centering
\includegraphics[width=10.5cm]{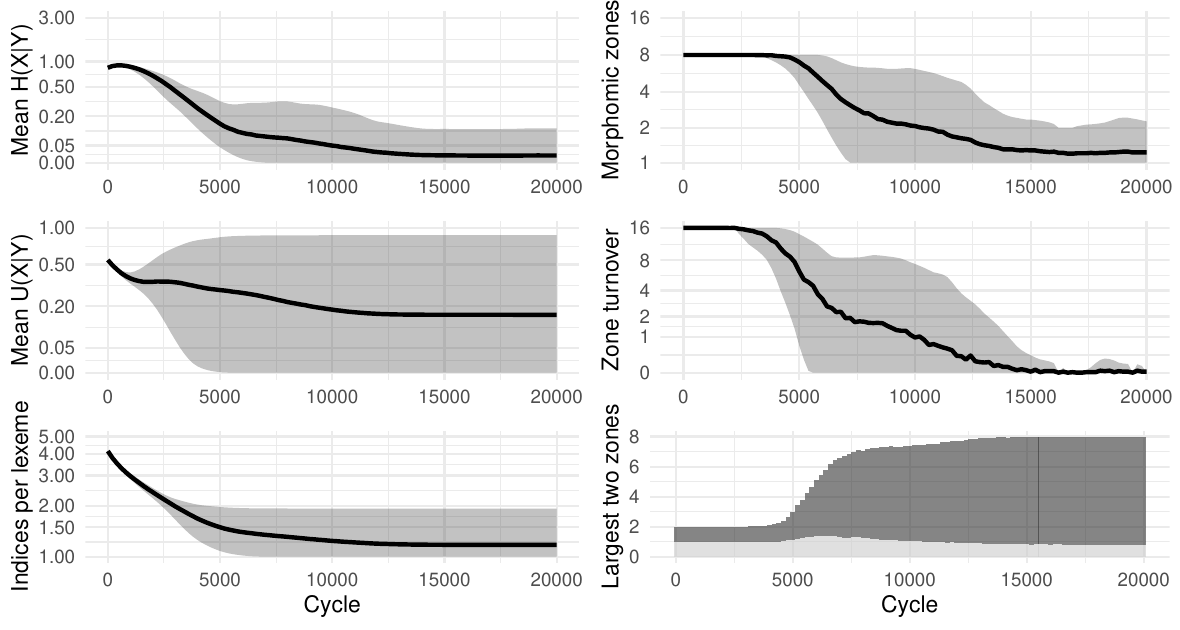}
\label{plot-meta-alpha05}
\end{figure}

\begin{figure}[ht!]
\caption{Sampling 10\% of pivot lexemes and 6 evidence cells from a uniform distribution, incorporating dissociative evidence ($\alpha = 1$). 100 simulations of 50,000 cycles.}
\centering
\includegraphics[width=10.5cm]{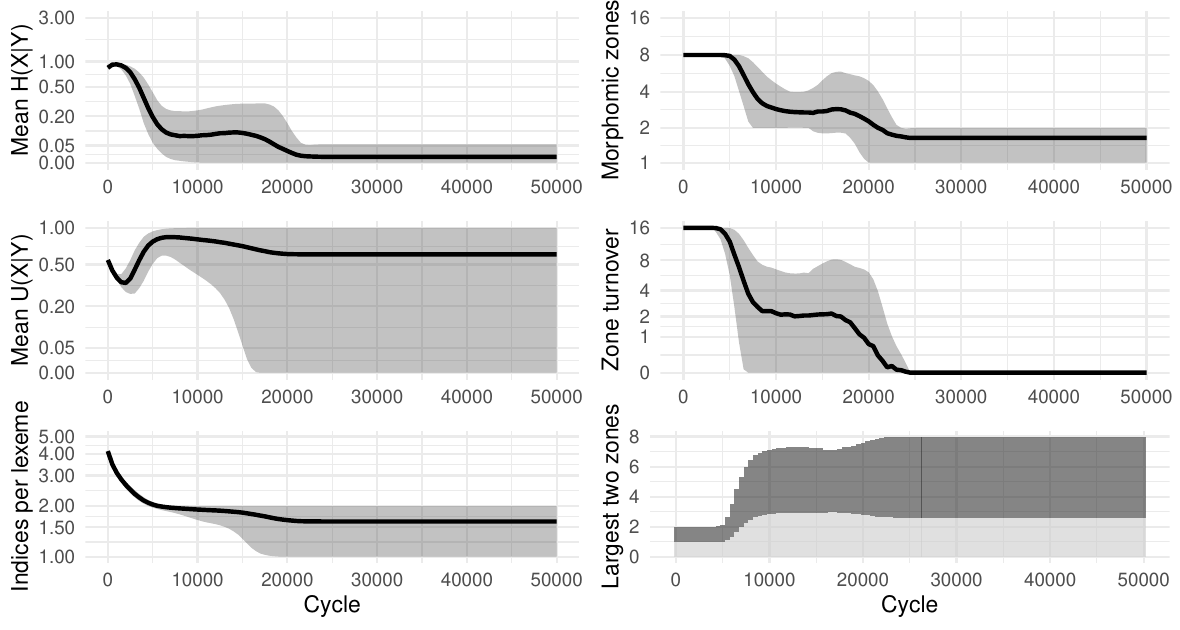}
\label{plot-meta-alpha1}
\end{figure}

\begin{figure}[ht]
\caption{Eight snapshots evenly spaced between cycle 0 (leftmost) and cycle 15,000 (rightmost) from one simulation incorporating dissociative evidence ($\alpha = 1$). Each snapshot shows 100 lexemes in rows, 8 cells in columns. Distinct allomorph indices in each lexeme are indicated by shading. Cells have been ordered horizontally so as to accentuate the two final morphomic zones most clearly.}
\centering
\includegraphics[width=12cm]{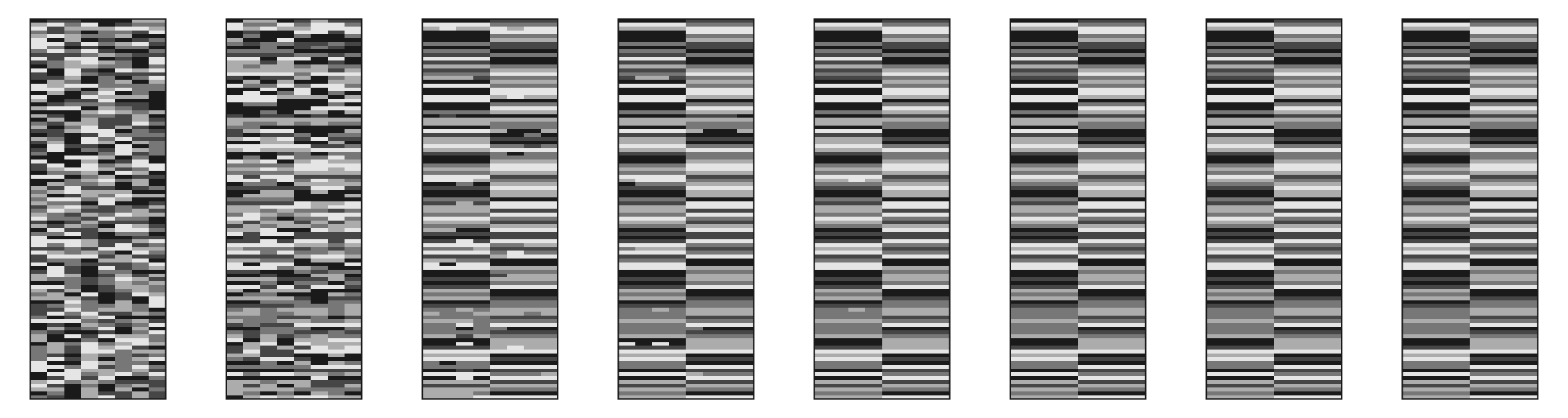}
\label{snapshots-appendix2}
\end{figure}

Figures \ref{plot-meta-alpha025}, \ref{plot-meta-alpha05} and \ref{plot-meta-alpha1} add dissociative evidence at strengths of $\alpha=0.25$, $\alpha=0.5$ and $\alpha=1$. As shown for rhizomorphomes (Section \ref{NE-model}), at the lowest values of $\alpha$, metamorphomes collapse, but at $\alpha = 0.5$ around half of the simulations reach a stable state with two metamorphomes by around 15,000 cycles, and at $\alpha = 1$, simulations attain a stable state by around 25,000 cycles, of which the majority retain two distinct morphomic zones. Thus at sufficient levels, dissociative evidence enables morphomic categories to emerge and persist. Figure~\ref{snapshots-appendix2} shows snapshots from within the first 15,000 cycles one of the simulations using $\alpha = 1$, in which two stable morphomic zones emerge.

\pagebreak




\end{appendices}


\FloatBarrier
\bibliography{sn-bibliography}% common bib file


% \includepdf[pages={1-7}]{Paper_code.pdf}

\end{document}